\pdfoutput=1

\documentclass[11pt]{article}

\usepackage{acl}

\usepackage{times}
\usepackage{latexsym}
\usepackage{soul}
\usepackage{url}
\usepackage{hyperref}
\usepackage{caption}
\usepackage{graphicx}
\usepackage{amsmath}
\usepackage{amsthm}
\usepackage{booktabs}
\usepackage{algorithm}
\usepackage{algorithmic}
\usepackage{bm}
\usepackage{amsfonts}
\usepackage{multirow}
\usepackage{makecell}
\usepackage{stfloats}
\usepackage{booktabs}
\usepackage{color}
\usepackage{graphicx}
\usepackage{subcaption} 
\usepackage{amssymb}

\usepackage[T1]{fontenc}

\usepackage[utf8]{inputenc}

\usepackage{microtype}

\usepackage{inconsolata}

\title{Fine-grainedly Synthesize Streaming Data Based On Large Language Models With Graph Structure Understanding For Data Sparsity} 

\author{Xin Zhang$^{\spadesuit}$\hspace{0.5mm}
Linhai Zhang$^{\spadesuit}$\hspace{0.5mm} 
Deyu Zhou$^{\spadesuit}$\thanks{~~Corresponding author.}\hspace{0.5mm} 
Guoqiang Xu$^{\diamondsuit}$\hspace{0.5mm}\\
        $^{\spadesuit}$\hspace{0.5mm}School of Computer Science and Engineering, Key Laboratory of Computer Network \\
and Information Integration, Ministry of Education, Southeast University, China 
\\ $^{\diamondsuit}$\hspace{0.5mm}SANY GROUP CO., LTD. \\
        \texttt{\{zhangxin, lzhang472, d.zhou\}@seu.edu.cn}\\
        \texttt{xuguoqiang-2012@hotmail.com}
        }

\begin{document}
\maketitle
\begin{abstract}
Due to the sparsity of user data, sentiment analysis on user reviews in e-commerce platforms often suffers from poor performance, especially when faced with extremely sparse user data or long-tail labels. Recently, the emergence of LLMs has introduced new solutions to such problems by leveraging graph structures to generate supplementary user profiles. However, previous approaches have not fully utilized the graph understanding capabilities of LLMs and have struggled to adapt to complex streaming data environments.
In this work, we propose a fine-grained streaming data synthesis framework that categorizes sparse users into three categories: Mid-tail, Long-tail, and Extreme. 
Specifically, we design LLMs to comprehensively understand three key graph elements in streaming data, including Local-global Graph Understanding, Second-Order Relationship Extraction, and Product Attribute Understanding, which enables the generation of high-quality synthetic data to effectively address sparsity across different categories.
Experimental results on three real datasets demonstrate significant performance improvements, with synthesized data contributing to MSE reductions of 45.85\%, 3.16\%, and 62.21\%, respectively.

\end{abstract}

\section{Introduction}

\begin{figure}[h] 
  \centering
  \includegraphics[width=0.48\textwidth]{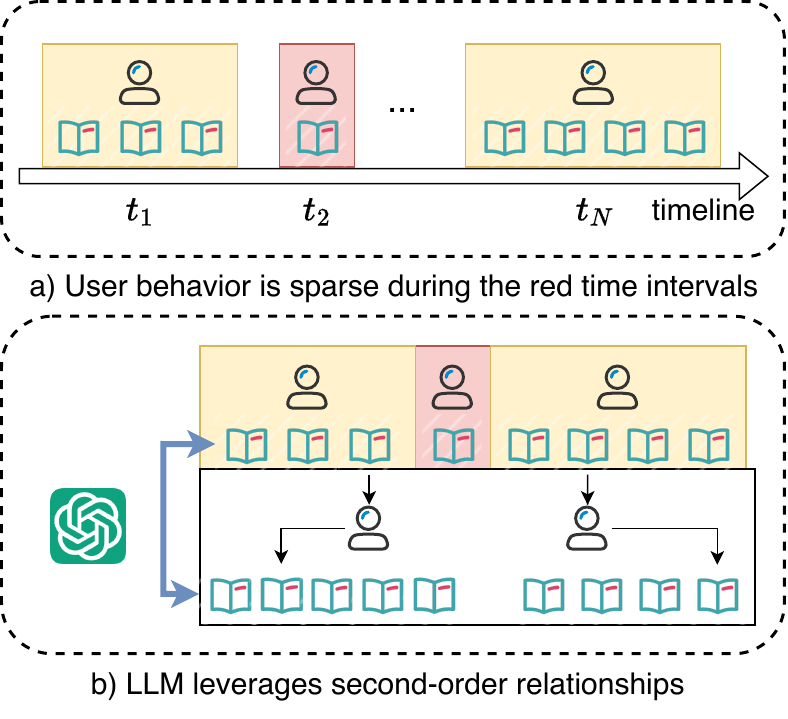}
    \caption{An example of temporal sparsity among users in streaming data. LLM leverages second-order relationships to synthesize similar product-user data, filling in the temporal gaps.}
    \label{fig:intro_midtail}
\end{figure}

Sentiment analysis for streaming users in E-commerce websites, as a form of dynamic sentiment analysis, holds significant importance and can be applied for various purposes such as personalized recommendations (\citealp{DBLP:conf/emnlp/ZhangZZ23,DBLP:conf/emnlp/WuSS023}). 
However, in the context of streaming data, user behavior on the timeline is often uneven, as illustrated in Figure~\ref{fig:intro_midtail}, with sparse behavior during certain time periods. This leads to data exhibiting non-uniform or sparse patterns and may result in issues such as cold starts or instability in the quality of learned representations by the model (\citealp{DBLP:conf/ijcai/Guo13,DBLP:conf/sigir/DuY00022}).
To address these challenges, previous methods for data sparsity have relied on supplementing graph information from the raw data (\citealp{DBLP:journals/corr/abs-2302-02151,DBLP:conf/sigir/WangZSWW23,DBLP:conf/sigir/ChenWHHXLH022}) or transferring knowledge from other datasets (\citealp{DBLP:conf/sigir/GaoWLWLYZL023,DBLP:conf/sigir/ZhuGZXXZL021}). Recently, meta-learning has also served as a popular solution for data sparsity (\citealp{DBLP:conf/sigir/WuZ23,DBLP:conf/kdd/Lu0S20,DBLP:conf/kdd/LeeIJCC19}). 
However, these methods face challenges due to the inherent sparsity of the dataset or difficulties in effectively transferring knowledge due to domain differences. 









Recently, large language models have emerged as an abstract form of large-scale knowledge graph, offering numerous new solutions for addressing the problem of data sparsity (\citealp{DBLP:conf/emnlp/LiZL023,DBLP:conf/emnlp/LeePSWJ23}). Some efforts are based on large language models' understanding of graph structure knowledge to solve sparsity issues, where first-order connectivity relationships are transformed into textual inputs for the large language model, aiming to achieve an initial understanding of the graph structure \cite{DBLP:journals/corr/abs-2311-00423}.
Additionally, there have been efforts to enhance user profiles by leveraging the social understanding capabilities of large language models and their grasp of anthropological knowledge, which have also made progress in addressing data sparsity \cite{DBLP:conf/emnlp/SunL0CAZJ23}.
However, these efforts either remain confined to understanding first-order relationships, failing to fully harness the potential of large language models in graph structure understanding, or solely focus on simple profile completion without adequately integrating graph structures, which all fail to cope with the evolving and more complex streaming data scenarios.

\begin{figure*}[!th]
\centering
\includegraphics[width=\textwidth]{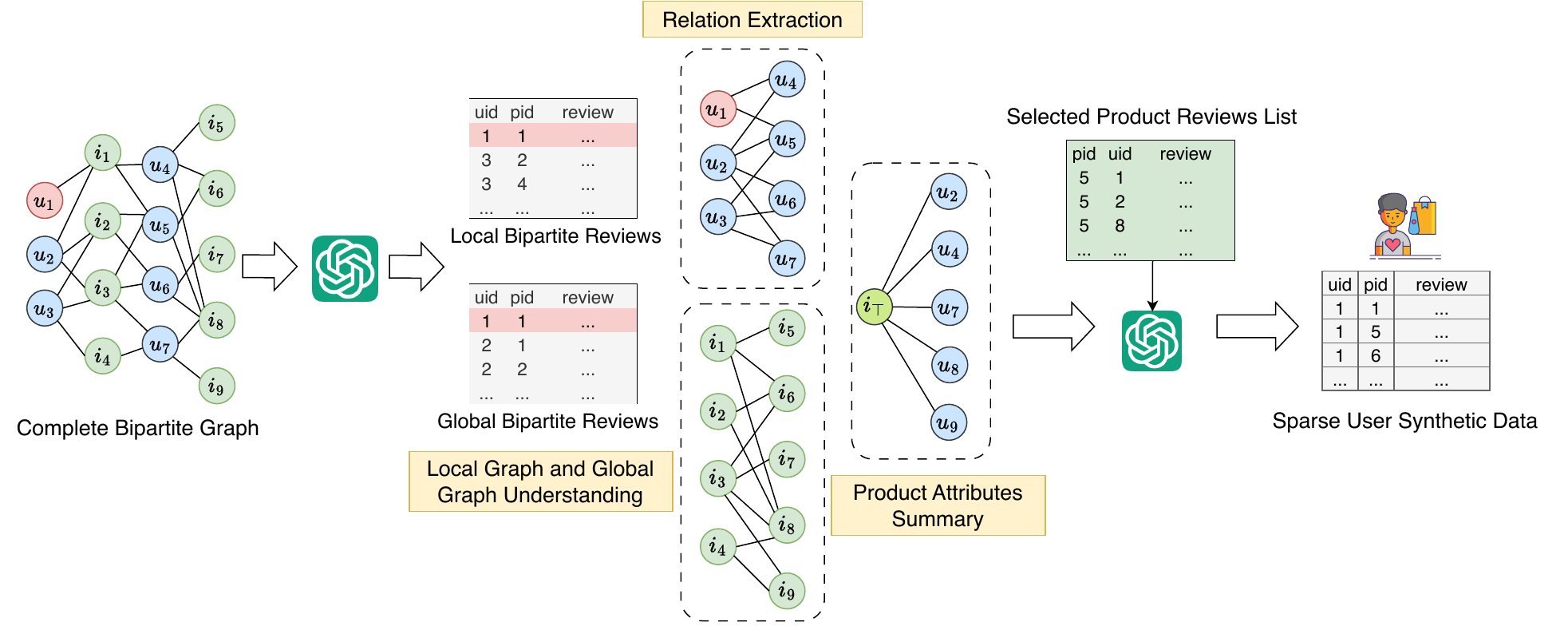}
\caption{Framework for utilizing LLM as a handler for streaming data sparsity. The bipartite graph stream serves as input; LLM needs to understand three key components in the graph: Local-Global Graph Understanding, Second-Order Relationship Extraction, and Product Attribute Understanding (where product information sometimes originates directly from the initial input and sometimes from other selected products under different rules); Finally, combining sparse user information with selected product information to obtain the final synthesized data, where the synthesized review data includes both review text and corresponding ratings.}
\label{fig:Framework}
\end{figure*}

Considering the temporal characteristics of streaming data and the spatial characteristics that evolve over time \cite{DBLP:conf/aaai/ParejaDCMSKKSL20,DBLP:conf/wsdm/SankarWGZY20, DBLP:conf/sigir/0001GRTY20}, we believe that, in addition to first-order relationships, taking into account second-order or even third-order relationships in the streaming graph is crucial for supplementing sparse user information. Furthermore, compared to first-order heterogeneous relationships, the homogeneity performance of second-order bipartite graphs in streaming data can better assist models in handling sparsity \cite{DBLP:conf/kdd/JiFJZT020}, highlighting the importance of including higher-order relationships. Additionally, with the introduction of the time dimension, the sparse behavior of users becomes more complex compared to static situations. For some users, their sparse issues are not caused by data sparsity itself. For example, in Figure~\ref{fig:intro_midtail}, certain users have sparse data due to temporal sparsity. Therefore, it is necessary to classify users based on various sparse categories and design solutions accordingly.

Based on these findings, we propose a fine-grained data synthesis framework that integrates LLM's comprehensive understanding of streaming graph structures and its comprehension of human sociological knowledge, aiming to address the data sparsity issue in streaming data. On one hand, considering the structure of streaming graphs, as illustrated in Figure~\ref{fig:Framework}, we incorporate three key elements into the framework to extract and maximize the utilization of streaming graph structural information for LLM. These elements include local-global graph understanding, user/product second-order relationship extraction, and product attribute understanding. On the other hand, considering the scarcity of users across different categories, users may be scarce in quantity or exhibit temporal imbalances. We categorize users into three types for investigation: mid-tail users (not scarce in quantity but scarce or imbalanced in the temporal dimension), long-tail users (scarce in quantity but not in spatial distribution), and extreme situation users (scarce in spatial dimension with few neighbors). LLM needs to extract effective streaming graph knowledge for these three types of users to complete data synthesis and supplement sparse data.
Our method demonstrates effectiveness across three real sparse datasets from Amazon.

\section{Related Work}

\subsection{User Data Sparsity}
Previous research has investigated two common scenarios regarding the availability of interaction information for sparse users: zero-shot and few-shot. In the zero-shot scenario, strategies involve leveraging auxiliary information or incorporating user attributes into preference representations to improve recommendation performance. For example, DropoutNet \cite{DBLP:conf/nips/VolkovsYP17} and Heater \cite{DBLP:conf/sigir/ZhuSSC20} adopt techniques like dropout strategies and pretrained collaborative filtering representations, respectively. Social networks are also used to enrich user representations \cite{DBLP:conf/recsys/SedhainSBXC14,DBLP:conf/sigir/DuY00022}, and cross-domain recommendation methods are effective in transferring preferences across domains \cite{DBLP:conf/cikm/HuZY18,DBLP:conf/wsdm/0008T20,DBLP:conf/sigir/GaoWLWLYZL023,DBLP:conf/sigir/ZhuGZXXZL021}. In the few-shot scenario, approaches focus on expanding potential interests beyond sparse interactions by leveraging semantic product associations, often extracted from graph-structured data \cite{DBLP:conf/cikm/WangZWZLXG18,DBLP:journals/corr/abs-2302-02151,DBLP:conf/sigir/WangZSWW23,DBLP:conf/sigir/ChenWHHXLH022}. 
Recently, meta-learning has also gained popularity as a solution for data sparsity (\citealp{DBLP:conf/sigir/WuZ23,DBLP:conf/kdd/Lu0S20,DBLP:conf/kdd/LeeIJCC19}). 
However, these methods face challenges due to the inherent sparsity of the dataset or difficulties in effectively transferring knowledge due to domain differences.
Moreover, all the aforementioned approaches overlook temporal information, addressing data sparsity solely from a static perspective, which cannot handle the sparsity issues in dynamic streaming data from real-world e-commerce platforms.

\subsection{LLMs as Data Annotator}

Large language models (LLMs) are widely used in data annotation due to their strong reasoning capabilities and vast knowledge. Research in this area mainly focuses on designing prompts to query LLMs or enhance their reasoning abilities. For instance, \citet{yu2024metamath} generates mathematical question answering data by rephrasing questions from different angles, while \citet{liang-etal-2023-prompting} uses Chain-of-Thought prompting for complex reasoning. \citet{ye-etal-2023-generating} leverage LLMs' coding generation abilities to create symbolic language data. Other studies explore new annotation tasks with LLMs, such as inferring user privacy \citep{staab2024beyond}, allocating annotation tasks between humans and LLMs \citep{li-etal-2023-coannotating}, and optimizing prompts for LLMs against distribution shifts \citep{li-etal-2023-robust}.
Recently, there have been some efforts specifically aimed at supplementing and expanding user data, focusing on the understanding of first-order neighbor information \cite{DBLP:journals/corr/abs-2311-00423} or user profiles \cite{DBLP:conf/emnlp/SunL0CAZJ23}.
However, these data supplementation efforts either overlook graph structures or underutilize graph structural information, failing to maximize the potential of LLM's understanding of graph structures.


\section{LLM as a Handler for Streaming Data Sparsity}
\subsection{Sparsity Handler Framework}
The streaming user-product graph in e-commerce platforms exhibits a tree-like structure with streaming characteristics \cite{DBLP:conf/sigir/Wang0WFC19}. To address the sparsity issue inherent in such data, we propose a novel fine-grained framework aimed at achieving maximal and effective exploration of user interests and synthesizing data through LLM's comprehensive understanding of all graph structural relationships in the streaming graph. Considering various sparse user scenarios, we categorize users into three types: mid-tail users (not sparse in quantity but sparse or imbalanced in the temporal dimension), long-tail users (sparse in quantity but not sparse in spatial distribution), and extreme situation users (sparse in spatial dimension with few neighbors). LLM needs to understand the following three types of graph structural elements and design solutions to generate synthetic data for each of these user categories accordingly.
\begin{itemize}
    \item \textbf{Local-Global Graph Understanding}: When building graphs based on streaming data, we divide them into different snapshots based on different time periods. In this paper, each of these snapshots is called a local graph. At the same time, there is a complete graph over the entire timeline, gradually getting bigger as time goes on, and we call this the global graph \cite{DBLP:conf/emnlp/JinQJR20}. For long-tail users, LLM needs to understand both of these graphs simultaneously to make the most of the knowledge of the graph structure.
    \item \textbf{Second-Order Relationship Extraction}: In contrast to traditional first-order relationship extraction, our emphasis lies in the extraction of second-order relationships. Such design stems from the bipartite nature of e-commerce data, where single-hop features may inadequately capture the relationships between nodes, whereas second-order relationships are crucial for enhancing the understanding of sparse user interests. In this paper, we specifically explore two types of second-order relationships: user-second-order relationships and product-second-order relationships.
    \item \textbf{Product Attribute Understanding}: To generate synthetic data, it's important for LLM to be able to use the original reviews about a product or combine the second-order homogenous relationships related to the product, which allows LLM to provide relevant summaries for the selected product attributes.
\end{itemize}



\subsection{LLM as Mid-tail Sparsity Handler}
In this paper, we introduce the concept of mid-tail users—individuals who contribute reviews within specific time frames but demonstrate varying behavior across different intervals, as is shown in  Figure~\ref{fig:intro_midtail}. 
These users exhibit moderate preferences and engagement levels, positioning themselves between the extensively studied realms of frequent engagement and the long tail. 
To enhance behavior analysis within this user category, our focus centers on improving the stability and quality of the model's learned representations across diverse time intervals.

\noindent\textbf{User Review Understanding. }For Mid-tail users, given their relatively abundant reviews, we directly generate user profiles using a subset of their own reviews. We randomly select $K$ reviews to input into LLM for user profile generation.
\begin{equation}
\text{User}_M = \mathbf{L L M} ({\mathrm{R_{chosen}} (u_{m})} ),\mathrm{P}_{p_{um}})
\end{equation}
where $\mathrm{R_{chosen}}(u_{m})$ represents the reviews selected for generating the profile of mid-tail user $u_{m}$, $\mathrm{P}_{p_{um}}$ is the prompt used for generating the profile of mid-tail users, and $\text{User}_M$ refers to the profiles generated for mid-tail users.

\begin{figure*}[!th]
\centering
\includegraphics[width=\textwidth]{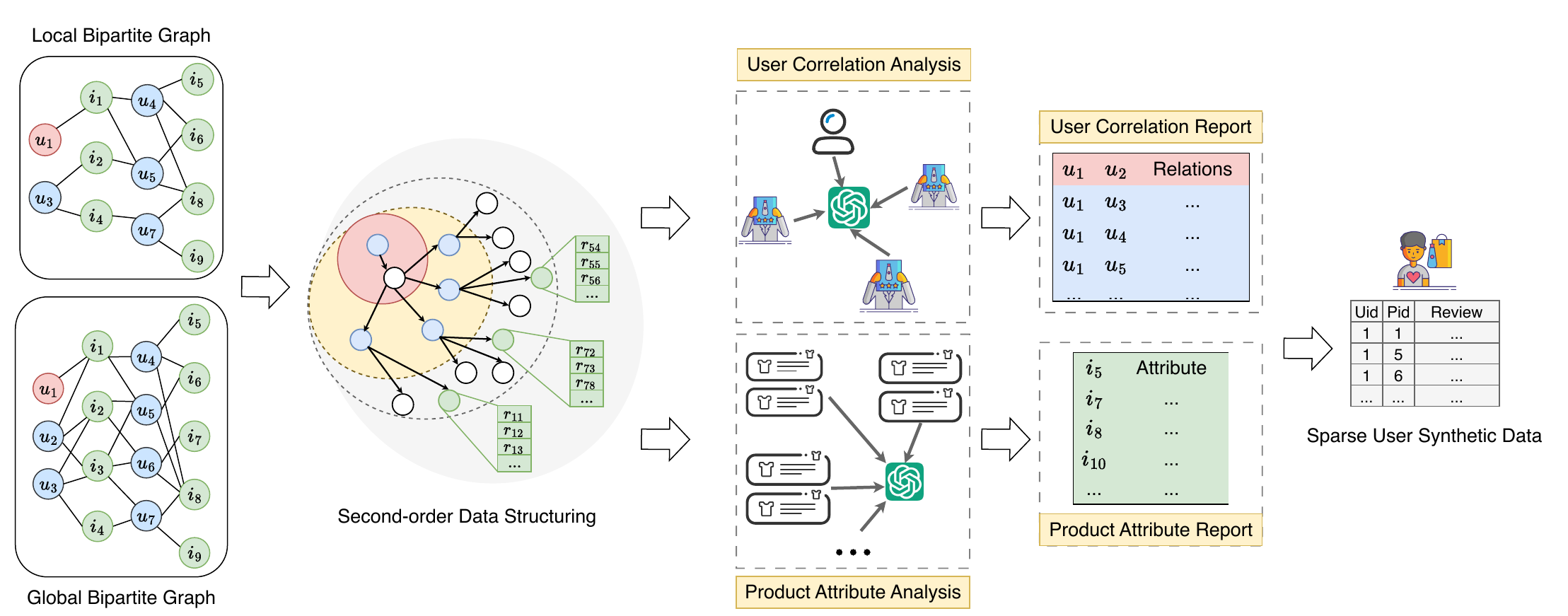}
\caption{Long Tail User Scenario. Local bipartite graphs and global bipartite graphs serve as inputs. LLM needs to simultaneously analyze the second-order homogeneous user relationships in both the local bipartite graph and the global bipartite graph of Long Tail Users to obtain supplementary Long Tail User profiles. It also needs to analyze the third-order product relationships corresponding to Long Tail Users in the global bipartite graph to obtain product profiles for data synthesis.}
\label{fig:longtail}
\end{figure*}

\noindent\textbf{Product Second-order Relationships. }The second-order homogeneous products interacted by users, given their similarity to the first-order products of users, serve as the product-side information for synthesized data here. Based on the first-order product relationships corresponding to user $u_{m}$, we extract the second-order homogeneous relationships associated with these products, forming a set as $\text{Second\_Order}(p_i) = \{(p_i, p_j), (p_i, p_k), \ldots\}$, which is also denoted as $\text{Second\_Order} (\text{First\_Order}(u_{m}))$.

Then, we randomly select $N$ products from $\text{Second\_Order}(p_i)$ and randomly choose five reviews from their corresponding reviews. These reviews were input into LLM to obtain their profiles, denoted as $\mathrm{P_{profile\_set}}(u_{m})$. 
\begin{equation}
\begin{split}
\mathrm{P_{profile\_set}}(u_{m}) = \mathbf{L} \mathbf{L M}(\mathrm{P}_{pm}, 
\text{Second\_Order} \\
(\text{First\_Order}(u_{m})))
\end{split}
\end{equation}
where $\mathrm{P}_{pm}$ is the prompt for generating the profile of products in the mid-tail scenario.

Subsequently, by utilizing LLM to understand the relationship between the original product profile and the second-order homogeneous product profiles, we selected a suitable list of products for synthetic data, formalized as $\mathrm{P_{set}}(u_{m} ) = \{p_j, p_k, \ldots \}$.
\begin{equation}
\begin{split}
\mathrm{P_{set}}(u_{m} ) = \mathbf{L} \mathbf{L M}(\mathrm{P}_{so}, \mathrm{P_{profile\_set}}(u_{m}), \\
\mathrm{P_{profile}}(\text{First\_Order}(u_{m})))
\end{split}
\end{equation}
where $\mathrm{P}_{so}$ is the prompt for identifying suitable second-order relationships.

Finally, retrieve the profile of the selected product for use in the subsequent data synthesis.
\begin{equation}
\begin{split}
\text{Product}_M = \mathrm{P_{profile\_set}}(u_{m})(\mathrm{P_{set}}(u_{m} ))
\end{split}
\end{equation}


\noindent\textbf{Mid-tail Data Synthesis. }We input the user profile and product profile into LLM to obtain the final synthesized data.
\begin{equation}
\text{Synthetic\_Data}_M = \mathbf{LLM}(\mathrm{P}_{sd}, \text{User}_M, \text{Product}_M)
\end{equation}
where $\mathrm{P}_{sd}$ is the prompt for synthetic data generation.

\subsection{LLM as Long-tail Sparsity Handler}

Long-tail users are defined as those who have only posted a small number of reviews, for example, once or twice. Such behavioral pattern poses challenges for modeling and implementing personalized analysis for them because predicting and capturing the interests and activity levels of such users is difficult \cite{DBLP:journals/tkde/LiLHS21}. Therefore, additional knowledge, such as second-order information, is needed to complement their profiles. 
By introducing LLM for semantic understanding of interest preferences, we can more effectively extract valuable information from second-order neighbor relationships. Meanwhile, the impact of user neighborhood graphs on preferences varies across different time periods. Incorporating the temporal influence, we design both long-term and short-term neighbor graphs to complement user information. 

As shown in Figure~\ref{fig:longtail}, the steps for synthesizing data for long-tail users are as follows. Firstly, we input the local and global bipartite graphs into LLM. Initially, LLM needs to mine user interests based on their own reviews, followed by supplementing user profiles through long-term and short-term second-order homogeneous relationships. Next, appropriate product profiles are generated by selecting from the second-order homogeneous neighbors of products. Finally, the data synthesis process is also completed by LLM.

\noindent\textbf{Local and Global Graphs. }We adopt the concept of discrete dynamic graphs, referring to previous definitions of dynamic temporal graphs in the analysis of streaming data \cite{DBLP:journals/corr/abs-2203-10480}. Specifically, the definition is as follows: a dynamic graph \( G_T = O_T \) for a time span \( T = [t_1 : t_n] \) is considered a Discrete Temporal Dynamic Graph (DTDG). Each stored observation \( o_{t_i} \) in \( O_T \) represents a snapshot of the graph \( o_{t_i} = (V_{t_i}, E_{t_i}, X_{t_i}) \), where \( V_{t_i} \), \( E_{t_i} \), and \( X_{t_i} \) denote nodes, edges, and the node features matrix observed at time \( t_i \).

\noindent\textbf{Local \& Global User Second-order Relationships. }For the local and global graphs, we extract the second-order homogeneous relationships of \( u_l \) as \( \text{Second\_Order\_Local}(u_l) \) and \( \text{Second\_Order\_Global}(u_l) \) respectively. These two sets of second-order homogeneous reviews, along with all reviews made by \( u_l \), are then input into LLM to generate the profile of \( u_l \).
\begin{equation}
\begin{split}
\text{User}_L = \mathbf{L L M} (\text{\text{Second\_Order\_Local} }(u_l),  \\
\text{Second\_Order\_Global}(u_l), \mathrm{R_{chosen}}(u_{l}), \mathrm{P}_{p_{ul}})
\end{split}
\end{equation}
Where $\mathrm{P}_{p_{ul}}$ is the prompt used for generating user profiles in the long-tail scenario.

\noindent\textbf{Global Product Second-order relationships. }Extracting user profiles in the long-tail scenario follows a process similar to that in the mid-tail scenario. Initially, product profiles are obtained through LLM, which is denoted as $\mathrm{P_{profile\_set}}(u_{l})$. Subsequently, a product list $\mathrm{P_{set}}(u_{l} )$ is selected for data synthesis by understanding the relationships between the original product and its second-order products. Finally, the profile information of the corresponding products is retrieved to prepare for the next step of data synthesis.
\begin{equation}
\begin{split}
\mathrm{P_{profile\_set}}(u_{l}) = \mathbf{L} \mathbf{L M}(\mathrm{P}_{pl}, 
\text{Second\_Order} \\
(\text{First\_Order}(u_{l})))
\end{split}
\end{equation}
\begin{equation}
\begin{split}
\mathrm{P_{set}}(u_{l} ) = \mathbf{L} \mathbf{L M}(\mathrm{P}_{so}, \mathrm{P_{profile\_set}}(u_{l}), \\
\mathrm{P_{profile}}(\text{First\_Order}(u_{l})))
\end{split}
\end{equation}
\begin{equation}
\begin{split}
\text{Product}_L = \mathrm{P_{profile\_set}}(u_{l})(\mathrm{P_{set}}(u_{l} ))
\end{split}
\end{equation}
where $\mathrm{P}_{pl}=\mathrm{P}_{pm}$ is the prompt for generating the profile of products in the long-tail scenario, and $\mathrm{P}_{so}$ is the same prompt as in the mid-tail scenario for identifying suitable second-order relationships.

\noindent\textbf{Long-tail Data Synthesis. }Finally, synthesized data is obtained by inputting both user profiles and product profiles into the LLM.
\begin{equation}
\text{Synthetic\_Data}_L = \mathbf{LLM}(\mathrm{P}_{sd}, \text{User}_L, \text{Product}_L)
\end{equation}
where $\mathrm{P}_{sd}$ is the same prompt as in the mid-tail scenario for synthetic data generation.

\subsection{LLM as Extreme Sparisity Handler}
For extreme cases, such as situations where users exhibit extreme sparsity, with not only their own reviews being sparse but also their surrounding neighbors being extremely sparse or even nonexistent, we propose using highly rated "popular" or popular products to construct pseudo data. 
With this approach, we ensure that the constructed data maintains high-quality information on the product side.  Importantly, this method maximizes the benefits of user representation learning while minimizing the loss generated by disrupting the graph structure.


\begin{figure}[h] 
  \centering
  \includegraphics[width=0.48\textwidth]{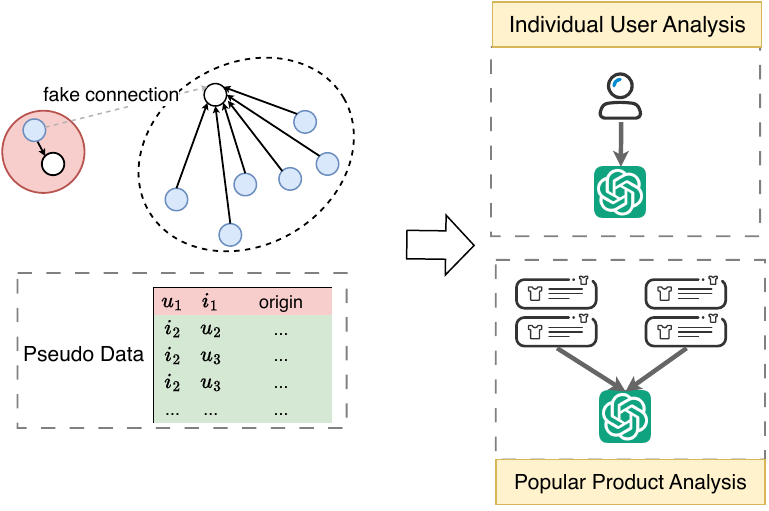}
    \caption{Extremely Sparse Scenario. Generating synthetic data by creating fake connections between the top products and Extreme Users to simulate pseudo interactions.}
    \label{fig:extreme}
\end{figure}

\begin{table*}[t]
    \centering
    \resizebox{1.8\columnwidth}{!}{
    \begin{tabular}{|l|c|c|c|c|c|c|}
    \hline
    \textbf{Dataset} & \textbf{Total num} & \textbf{Avg r/u} & \textbf{Avg r/p}  & \textbf{Sparse r/u} & \textbf{Long Tail r/u} & \textbf{Avg so/u} \\
    \hline
    Magazine\_Subscriptions & 2330 & 6.70 & 14.84 & 5.19 & 30.00 & 166.61 \\
    Appliances & 203 & 4.32 & 4.23  & 2.00 & 7.50 & 18.64 \\
    Gift\_Cards & 2966 & 6.49 & 20.04  & 5.35 & 30.00 & 242.08 \\
    \hline
    \end{tabular}}
\caption{Statistical information of sparse Amazon datasets. 'Avg r/u' means the average associated reviews number for users, and 'Avg r/p' means the average associated reviews number for products. 
'Avg so/u' means the average number of second-order homogeneous neighbors per user.}
\label{tab:dataset_statistic}
\end{table*}

\noindent\textbf{User Profile Summarization and Top Product Choosing. }
Due to the scarcity of reviews and neighbors for such users, profiles can only be obtained from their own reviews. Subsequently, $M$ products are selected from the top products and paired with users to create pseudo connections. The profiles of the selected products are then generated using LLM and finally combined with user profiles to obtain synthetic data.
\begin{equation}
\text{User}_E = \mathbf{L L M} ( \mathrm{R}(u_{e}),\mathrm{P}_{p_{ue}})
\end{equation}
\begin{equation}
\text{Product}_E = \mathbf{L L M} ( {\mathrm{R_{chosen}} (Top_p)},\mathrm{P}_{p_{pe}})
\end{equation}
\begin{equation}
\text{Synthetic\_Data}_E = \mathbf{LLM}(\mathrm{P}_{sd}, \text{User}_E, \text{Product}_E)
\end{equation}
where $Top_p$ refers to the selected popular product, $\mathrm{P}_{p_{ue}}$ is the prompt for generating the profile of extreme scenario users, and $\mathrm{P}_{p_{pe}}$ is the prompt for generating the profile of extreme scenario products.

\begin{table*}[htbp]
    \centering
    \resizebox{2\columnwidth}{!}{
\begin{tabular}{|l|c|c|c|c|c|c|c|c|}
\hline
\textbf{Category} & \textbf{Normal}& \textbf{Proportion} & \textbf{Mid-tail}& \textbf{Proportion} & \textbf{Long-tail}& \textbf{Proportion} & \textbf{Extreme}& \textbf{Proportion} \\ \hline
Magazine\_Subscriptions & 183 & 52.59\% & 8 & 2.30\% & 154 & 44.25\% & 3 & 0.86\% \\
Appliances & 11 & 23.40\% & 2 & 4.26\% & 19 & 40.43\% & 15 & 31.91\% \\
Gift\_Cards & 203 & 44.42\% & 45 & 9.85\% & 209 & 45.73\% & 0 & 0.00\% \\ \hline
    \end{tabular}}
\caption{Statistics of the number of users at different levels of sparsity.}
\label{tab:sparse_user_proportion}
\end{table*}

\begin{table*}[t]
\centering
\resizebox{2\columnwidth}{!}{
\begin{tabular}{|c|l|ccccccc|}
\hline
\multirow{2}{*}{\textbf{Dataset}} & \multicolumn{1}{c|}{\multirow{2}{*}{\textbf{Method}}} & \multicolumn{7}{c|}{\textbf{Criteria}}                                         \\ \cline{3-9} 
                                  & \multicolumn{1}{c|}{}                                 & Accuracy($\uparrow$)     & Precision($\uparrow$)     & Recall($\uparrow$)     & F1($\uparrow$)     & MSE($\downarrow$)    & RMSE($\downarrow$)     & MAE($\downarrow$)       \\ \hline
\multirow{7}{*}{\textit{Magazine\_Subscriptions}}         
         & BiLSTM+Att & 0.6910 & 0.4040 & 0.4054 & 0.4019 & 1.5021 & 1.2256 & 0.5837 \\
         & Bert-Sequence & 0.6953 & 0.2589 & 0.3049 & 0.2791 & 1.2918 & 1.1366 & 0.5451 \\
         & NGSAM & - & - & - & - & 1.1929 & 1.0922 & 0.7385 \\

         & CHIM & 0.7143 & 0.2381 & 0.3333 & 0.2778 & 1.0476 & 1.0235 & 0.4762 \\
         & IUPC & 0.7039 & 0.2671 & 0.3499 & 0.3016 & 0.9442 & 0.9717 & 0.4635 \\
         & \textbf{DC-DGNN} & 0.7554 & 0.4290 & 0.4107 & 0.4016 & 0.7768 & 0.8814 & 0.3820 \\
         & \textbf{DC-DGNN}$^{*}$ & \textbf{0.7983} & \textbf{0.6879} & \textbf{0.5853} & \textbf{0.5385} & \textbf{0.4206} & \textbf{0.6485} & \textbf{0.2575} \\

         \hline
\multirow{7}{*}{\textit{\makecell{Appliances}}}         
         & BiLSTM+Att & 0.7143 & 0.2381 & 0.3333 & 0.2778 & 1.0476 & 1.0235 & 0.4762 \\
         & Bert-Sequence & 0.7143 & 0.2381 & 0.3333 & 0.2778 & 1.0476 & 1.0235 & 0.4762 \\
         & NGSAM & - & - & - & - &0.6885 & 0.8298 & 0.5693 \\ 

         & CHIM & 0.6905 & 0.4392 & 0.3109 & 0.3341 & 0.9967 & 0.9983 & 0.4742   \\ 
         & IUPC & 0.7143 & 0.2381 & 0.3333 & 0.2778 & 1.0476 & 1.0235 & 0.4762 \\
         & \textbf{DC-DGNN} & 0.7143 & 0.2381 & 0.3333 & 0.2778 & 1.0476 & 1.0235 & 0.4762 \\
         & \textbf{DC-DGNN}$^{*}$ & \textbf{0.8571} & \textbf{0.5441} & \textbf{0.5833} & \textbf{0.5625} & \textbf{0.6667} & \textbf{0.8165} & \textbf{0.2857} \\

         \hline
\multirow{7}{*}{\textit{\makecell{Gift\_Cards}}}       
         & BiLSTM+Att & 0.8788 & 0.4696 & 0.2586 & 0.2505 & 0.3030 & 0.5505 & 0.1684 \\
         
         & Bert-Sequence & 0.8754 & 0.2189 & 0.2500 & 0.2334 & 0.3064 & 0.5535 & 0.1717 \\
         & NGSAM & - & - & - & - &0.2494 & 0.4994 & 0.2949 \\

         & CHIM & 0.8754 & 0.2189 & 0.2500 & 0.2334 & 0.3064 & 0.5535 & 0.1717 \\
         & IUPC & 0.8754 & 0.2189 & 0.2500 & 0.2334 & 0.3064 & 0.5535 & 0.1717 \\
         & \textbf{DC-DGNN} & 0.8754 & 0.2189 & 0.2500 & 0.2334 & 0.3064 & 0.5535 & 0.1717 \\
         & \textbf{DC-DGNN}$^{*}$ & \textbf{0.8956} & \textbf{0.4384} & \textbf{0.5000} & \textbf{0.4671} & \textbf{0.1145} & \textbf{0.3383} & \textbf{0.1077} \\

         \hline
\end{tabular}
}
\caption{Results of sentiment analysis on streaming user reviews across three real-world Amazon datasets. $\downarrow$ indicates the smaller the metrics, the better the method, while $\uparrow$ indicates the larger the metrics, the better the method. The score marked as bold means the best performance among all the methods.}
\label{tab:main_result}
\end{table*}

\subsection{Streaming Synthetic Data Validation Task}
Our synthesizing framework for handling sparse user data is validated in the context of sentiment analysis for streaming user reviews. Within the domain of sentiment analysis applied to streaming user reviews, the reviews are organized chronologically as $  E=\bigl\{\mathcal{E}_1, \ldots, \mathcal{E}_T\bigl\} $. Each review $\mathcal{E}_i$ is represented as $(u_i, p_i, t_i, d_i)$, where $t_i$ denotes the timestamp of review $d_i$, $u_i$ represents the user who wrote the review $d_i$, and $p_i$ indicates the product being reviewed.

The objective of this task is to predict the user's rating $y$ towards the product under the current condition $\mathcal{E}_t$, utilizing historical information $ \bigl\{\mathcal{E}_1, \ldots, \mathcal{E}_{t-1}\bigl\} $, and to learn a mapping function between the user's rating $y$ and the condition $\mathcal{E}_t$, represented as $y=\operatorname{f}(\mathcal{E}_t|\bigl\{\mathcal{E}_1, \ldots, \mathcal{E}_{t-1}\bigl\})$.

\section{Experiments}

\subsection{Experiments Setup}

\textbf{Details of dataset information statistics and sparsity level dividing.} For quick validation of our method, we selected three datasets with the smallest data size from the Amazon dataset \cite{DBLP:conf/emnlp/NiLM19}, namely Magazine\_Subscriptions, Appliances, and Gift\_Cards. For these datasets, we retained the data in its original form without further cleaning to preserve the data in its most original state. Statistical analysis was conducted on various aspects of the datasets, and the results are presented in Table~\ref{tab:dataset_statistic}. Subsequently, based on the definitions of mid-tail users, long-tail users, and extreme situation users as outlined in this paper, we divided the users in the dataset into these three categories. The specific dividing process is illustrated in the appendix, and the numbers and proportions of users in each category after dividing are shown in Table~\ref{tab:sparse_user_proportion}.

\noindent\textbf{Baselines.} We selected two types of baseline models, including \textbf{\textit{Text-based model}}: BiLSTM+Att, Bert-Sequence \citep{DBLP:conf/naacl/DevlinCLT19}; and \textbf{\textit{User and Product-based model}}: CHIM \cite{DBLP:conf/emnlp/Amplayo19}, IUPC \citep{DBLP:conf/coling/LyuFG20}, NGSAM \cite{DBLP:conf/aaai/ZhouZZH21}, DC-DGNN \cite{DBLP:conf/emnlp/ZhangZZ23}. Among them, DC-DGNN is a continuous dynamic graph learning model specially designed for streaming data. DC-DGNN* refers to the results achieved by training on a combination of raw data and synthetic data corresponding to three categories of sparse users, and then testing on the original test data.

\noindent\textbf{Implementation details. }For user and product embeddings, all models are set to 128 dimensions. The batch size is 8, and the learning rate is 3e-5, with a total of 2 epochs. We formalize the prediction of sentiment analysis over time as a classification problem, and evaluate our model using the following seven metrics: Accuracy, Precision, Recall, F1-score, Mean Squared Error (MSE), Root Mean Squared Error (RMSE), and Mean Absolute Error (MAE). The training and test sets are split using a ratio of 9:1.

\begin{table*}[t]
\centering
\resizebox{2\columnwidth}{!}{
\begin{tabular}{|c|l|ccccccc|}
\hline
\multirow{2}{*}{\textbf{Dataset}} & \multicolumn{1}{c|}{\multirow{2}{*}{\textbf{Method}}} & \multicolumn{7}{c|}{\textbf{Criteria}}                                         \\ \cline{3-9} 
                                  & \multicolumn{1}{c|}{}                                 & Accuracy($\uparrow$)     & Precision($\uparrow$)     & Recall($\uparrow$)     & F1($\uparrow$)     & MSE($\downarrow$)    & RMSE($\downarrow$)     & MAE($\downarrow$)       \\ \hline
\multirow{4}{*}{\textit{Magazine\_Subscriptions}}         
         & DC-DGNN$^{*}$ & 0.7983 & 0.6879 & 0.5853 & 0.5385 & 0.4206 & 0.6485 & 0.2575 \\
         & DC-DGNN-M & \textbf{0.8155} & 0.6785 & \textbf{0.6044} & \textbf{0.6225} & 0.4292 & 0.6551 & 0.2489 \\
         & DC-DGNN-L & 0.7725 & \textbf{0.6826} & 0.5029 & 0.4661 & 0.5365 & 0.7324 & 0.2961 \\
         & DC-DGNN-E & 0.8112 & 0.5715 & 0.5521 & 0.5484 & \textbf{0.4077} & \textbf{0.6385} & \textbf{0.2446} \\
         \hline
\multirow{4}{*}{\textit{\makecell{Appliances}}}         
          & \textbf{DC-DGNN}$^{*}$ & \textbf{0.8571} & \textbf{0.5441} & \textbf{0.5833} & \textbf{0.5625} & \textbf{0.6667} & \textbf{0.8165} & \textbf{0.2857} \\
         & DC-DGNN-M & 0.7143 & 0.2381 & 0.3333 & 0.2778 & 1.0476 & 1.0235 & 0.4762 \\
         & DC-DGNN-L & 0.6667 & 0.2593 & 0.3111 & 0.2828 & 0.6190 & 0.7868 & 0.4286 \\
         & DC-DGNN-E & 0.7143 & 0.2381 & 0.3333 & 0.2778 & 1.0476 & 1.0235 & 0.4762 \\
         \hline
\multirow{3}{*}{\textit{\makecell{Gift\_Cards}}}       
         & \textbf{DC-DGNN}$^{*}$ & \textbf{0.8956} & \textbf{0.4384} & \textbf{0.5000} & \textbf{0.4671} & \textbf{0.1145} & \textbf{0.3383} & \textbf{0.1077} \\
         & DC-DGNN-M & 0.8754 & 0.2234 & 0.2500 & 0.2359 & 0.1448 & 0.3805 & 0.1313 \\
         & DC-DGNN-L & 0.8754 & 0.2189 & 0.2500 & 0.2334 & 0.3064 & 0.5535 & 0.1717 \\
         \hline
\end{tabular}
}
\caption{Results of interpolating sparse user data across different categories. {DC-DGNN}$^{*}$ represents the results obtained by synthesizing data from Mid-tail, Long-tail, and Extreme user categories. DC-DGNN-M refers to the results obtained by synthesizing data from only Mid-tail users, DC-DGNN-L from only Long-tail users, and DC-DGNN-E from only Extreme users.}
\label{tab:catagory_cmp}
\end{table*}

\subsection{Temporal Interpolation Strategy}
Firstly, we classify users into three categories based on the scheme outlined in the appendix, each category exhibiting varying levels of sparsity for different reasons. Then, using our interpolation position search method, we identify the positions where interpolation is required for each category of data. As for the design of the interpolation scheme, we split the entire dataset into 10 timespans. Within each timespan, we check for the presence of corresponding user data. In cases where data is missing, we apply the appropriate interpolation scheme.
After interpolation, we guarantee data availability for each time period and maintain a total data count exceeding 10 for each user. Following this method, we obtain the number of interpolations for each data category in each dataset, as presented in Table~\ref{tab:Interpolated}. The distribution of interpolation positions for all data types across all datasets over time intervals is illustrated in Figure~\ref{fig:interpolated_distribution} in the appendix.

\begin{table}[htbp]
\centering
\resizebox{1\columnwidth}{!}{
\begin{tabular}{l|c|c|c}
\hline
\textbf{Category} & \textbf{Mid-tail} & \textbf{Long-tail} & \textbf{Extreme} \\ \hline
Magazine\_Subscriptions & 67 & 1287 & 26 \\ \hline
Appliances & 15 & 158 & 126 \\ \hline
Gift\_Cards & 358 & 1753 & 0 \\ \hline
\end{tabular}}
\caption{Statistics of Interpolated Review Count.}
\label{tab:Interpolated}
\end{table}

\subsection{Main Results}
The main experimental results are shown in Table~\ref{tab:main_result}. Firstly, we focus on the information provided by the model performance without the inclusion of synthetic data. It is worth noting that when observing the results in Appliances and Gift\_Cards, we can see clear result repetitions. For example, in Appliances, the MAE performance of many models is 0.4762, while in Gift\_Cards, the performance of many models is 0.1717. On Gift\_Cards, even user-based models perform worse than BiLSTM+Att. The likely reason for this phenomenon is the lack of data or data quality issues, which can be considered as a manifestation of the cold start problem to some extent. However, as mentioned earlier, these are all real situations existing in the real dataset that we must address. 
Therefore, it is crucial to focus on whether synthetic data can address this issue. After incorporating synthetic data into DC-DGNN as DC-DGNN*, it can be observed that the performance of prediction has been significantly improved compared to before.
We achieved a considerable improvement of 45.85\%, 3.16\%, and 62.21\% in the MSE metric for the three datasets. This result not only demonstrates the effectiveness of our synthetic data strategy but also illustrates that even in the presence of significant quality issues in the dataset, our data synthesis framework is still able to cope well, generating effective data and rescuing the data from the "cold start" problem.

\subsection{Sparsity Resolver}
To validate the effectiveness of each proposed component, we conducted the Sparsity Resolver experiment to assess the efficiency of data synthesis for each category, denoted as -M, -L, and -E for mid-tail, long-tail, and extreme users, respectively. 
As shown in Table~\ref{tab:catagory_cmp}, we found that combining data from all three categories generally resulted in the best performance across various datasets, such as Appliances and Gift\_Cards. However, in some cases, using only one type of supplementation led to the optimal outcome, as observed in Magazine\_Subscriptions.
This is because the datasets considered in this study are small-scale datasets, and introducing more data could introduce additional noise, potentially leading to a decrease in predictive performance. 
It is worth noting that, there was no change in performance in the -M and -E cases of the Appliances dataset, likely due to the small number of synthetic data introduced. This is reasonable, as attempting to improve performance by introducing only a few data points, as shown in Table~\ref{tab:Interpolated}, is also unlikely. 
As for the -L case of the Gift\_Cards dataset, overfitting still occurred, likely due to the severe imbalance of the original labels in this dataset, with proportions corresponding to labels 5, 4, 3, 2, and 1 being [0.9258, 0.0519, 0.0111, 0.0074, 0.0037] respectively.
Introducing a large amount of similar data under the long-tail scenario exacerbated this imbalance. However, it is worth mentioning that the overfitting phenomenon during training on the Gift\_Cards dataset was mitigated to some extent when combining the synthesis data from all three categories.


\subsection{Vocabulary Richness Analysis} 
\begin{figure}[h] 
  \centering
  \includegraphics[width=0.48\textwidth]{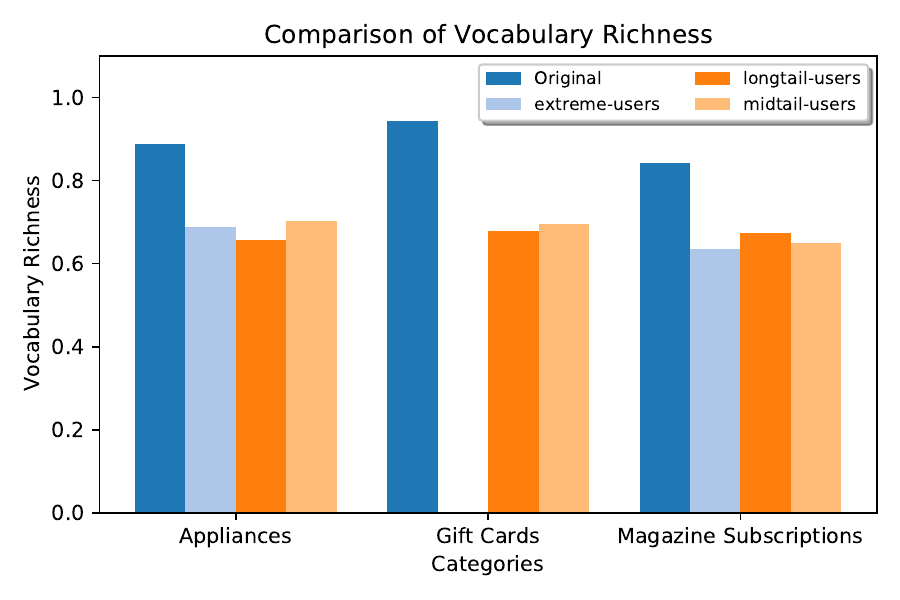}
    \caption{Vocabulary Richness Comparison.}
    \label{fig:vocabulary_richness}
\end{figure}

To assess the quality of the LLM synthetic data, we utilize NLTK\footnote{\url{https://www.nltk.org/}} to compute the overall average vocabulary richness of the synthesized data across different sparsity categories. We then compare these averages with those of the original data, as illustrated in Figure~\ref{fig:vocabulary_richness}. We observe that LLM exhibits results consistent with previous findings \cite{DBLP:conf/emnlp/LiZL023}, indicating a potential lack of diversity in the generated text. Across each category on the three datasets, the vocabulary richness of the text synthesized by LLM is lower than that of the original dataset and demonstrates a relatively consistent level of richness across each category of synthetic data.

\section{Conclusion}
In this paper, we address the challenge of data sparsity in sentiment analysis on streaming user reviews. We propose a fine-grained streaming data synthesis framework that categorizes sparse users into three categories. By designing LLM to understand various graph structures in streaming data, we generate high-quality synthetic data, effectively improving sentiment analysis performance. Experimental results demonstrate significant MSE reductions on three real datasets, highlighting the effectiveness of our approach in overcoming data sparsity challenges in e-commerce platforms.

\section*{Limitations}
Although our data synthesis approach has achieved excellent results in addressing user data sparsity, we still believe it has some limitations:
\begin{itemize}
    \item For the selection of next-hop neighbors, we adopt random sampling to save time costs. While this approach has little impact on cases with small sample sizes in this study, it may introduce noticeable biases in results when dealing with large sample sizes, as the randomness of sampling at different times becomes evident. To address this issue, we believe that future research can focus on designing more sophisticated and efficient selection schemes.
    \item 
    For the categorization of sparse data into different types, as we only examined small-scale datasets, the behavioral differences among users were not as apparent. In the future, investigating more diverse dataset types could help validate the effectiveness of the framework or reveal any shortcomings.
    \item 
    We do not further explore the ability of LLM to understand local and global graphs, nor explore the differences in understanding between the two. In fact, this is a topic worth investigating.
\end{itemize}

\bibliography{custom}
\clearpage
\appendix

\section{Experiment Details}
\label{sec:exp}

\noindent\textbf{Sparse user split details.} 
In the datasets considered in this paper, we preliminarily define users with more than 5 reviews as non-data-sparse users, and users with 5 or fewer reviews as data-sparse users based on clustering results. We further calculate the proportion of total reviews corresponding to these two categories of users in the dataset based on intervals of 0 to 5 and 5 to 10. Table~\ref{tab:dif_user_statistic} presents the statistical results. From the statistical results, it can be observed that reviews from these two categories of users already constitute the majority of reviews. Therefore, this paper only discusses these two categories. More specifically, users with 0 to 5 reviews are categorized into the data-sparse category (long-tail or extreme), while users with 5 to 10 reviews are categorized into the non-data-sparse category. For more detailed split rules, please refer to the corresponding introduction in the subsequent sections.

\begin{figure*}[htbp]
  \centering
  \begin{subfigure}[b]{0.41\textwidth}
    \includegraphics[width=\textwidth]{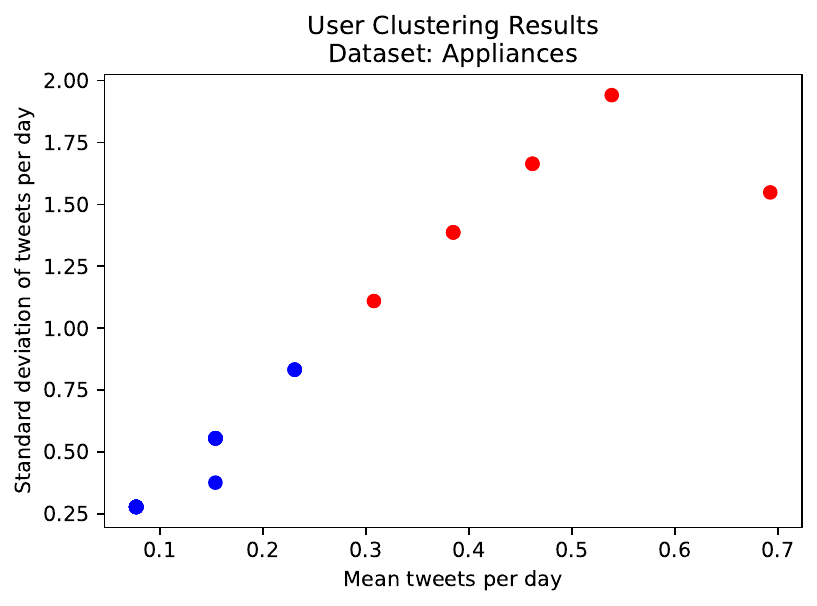}
    \caption{Appliances\_Mid-tail\_Split}
    \label{fig:sub1}
  \end{subfigure}
  \hfill
  \begin{subfigure}[b]{0.41\textwidth}
    \includegraphics[width=\textwidth]{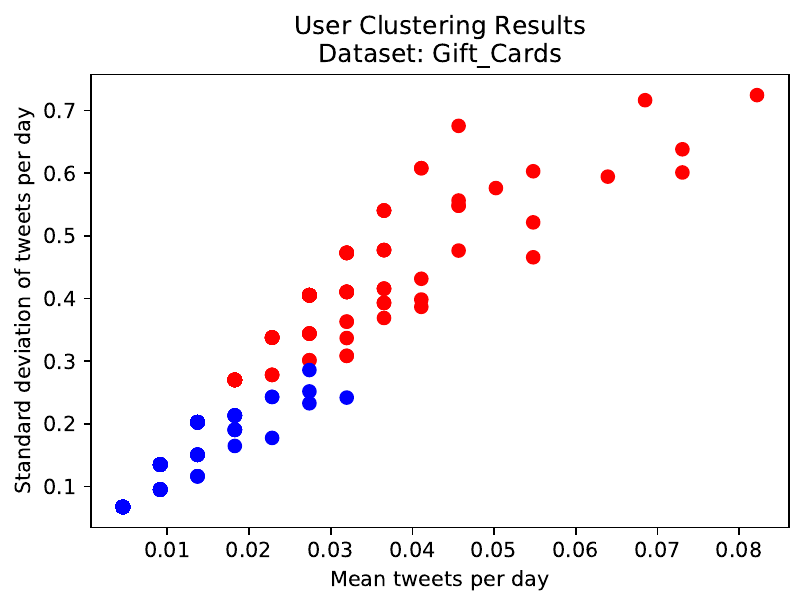}
    \caption{Gift\_Cards\_Mid-tail\_Split}
    \label{fig:sub4}
  \end{subfigure}
  \vskip\baselineskip 
  \begin{subfigure}[b]{0.41\textwidth}
    \includegraphics[width=\textwidth]{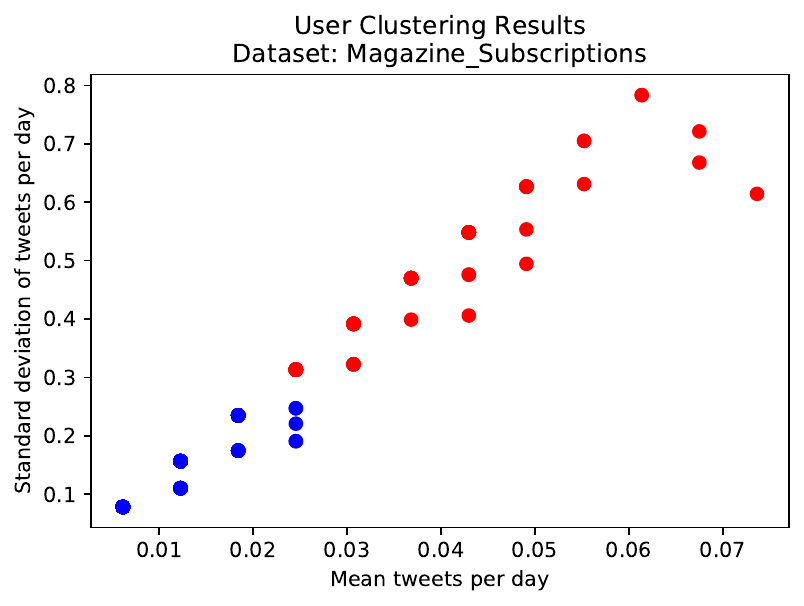}
    \caption{Magazine\_Subscriptions\_Mid-tail\_Split}
    \label{fig:sub5}
  \end{subfigure}
  \caption{Non-Data Sparse User Division. This section discusses users who are sparse in time rather than in data. The data points in the upper right corner indicate users with abundant but uneven data. The red dots in the figure are defined as mid-tail users.}
  \label{fig:nonsparse_division}
\end{figure*}

\noindent\textbf{Mid-tail user split.} 
Figure~\ref{fig:nonsparse_division} illustrates a further division of sparse users who are not data-scarce but sparse in the temporal. We first calculate the number of reviews for each user per day and then compute statistical indicators such as mean, standard deviation, minimum, and maximum review counts for each user. Subsequently, based on these statistical data, we apply the K-means algorithm to divide users into two groups. The meanings of the different sections in the figure are as follows:

\begin{itemize}
\item Top-right Users: These users have a higher average daily review count and exhibit greater variability in review counts. This may indicate highly active users whose review frequency fluctuates significantly, potentially influenced by external factors.

\item Top-left Users: These users also have a higher average daily review count, but with relatively lower variability. This suggests another group of highly active users whose review frequency remains more stable, and less influenced by external factors.

\item Bottom-right Users: Despite a lower average daily review count, these users display considerable variability in review counts. This might represent less active users who occasionally engage in bursts of reviewing but are generally less active.

\item Bottom-left Users: With a lower average daily review count and less variability, these users are likely less active overall and maintain a consistently low review frequency.

\end{itemize}

Among all these sections, we select Top-right Users and Bottom-right Users as \textit{Mid-tail users}.

\begin{figure*}[htbp]
  \centering
  \begin{subfigure}[b]{0.45\textwidth}
    \includegraphics[width=\textwidth]{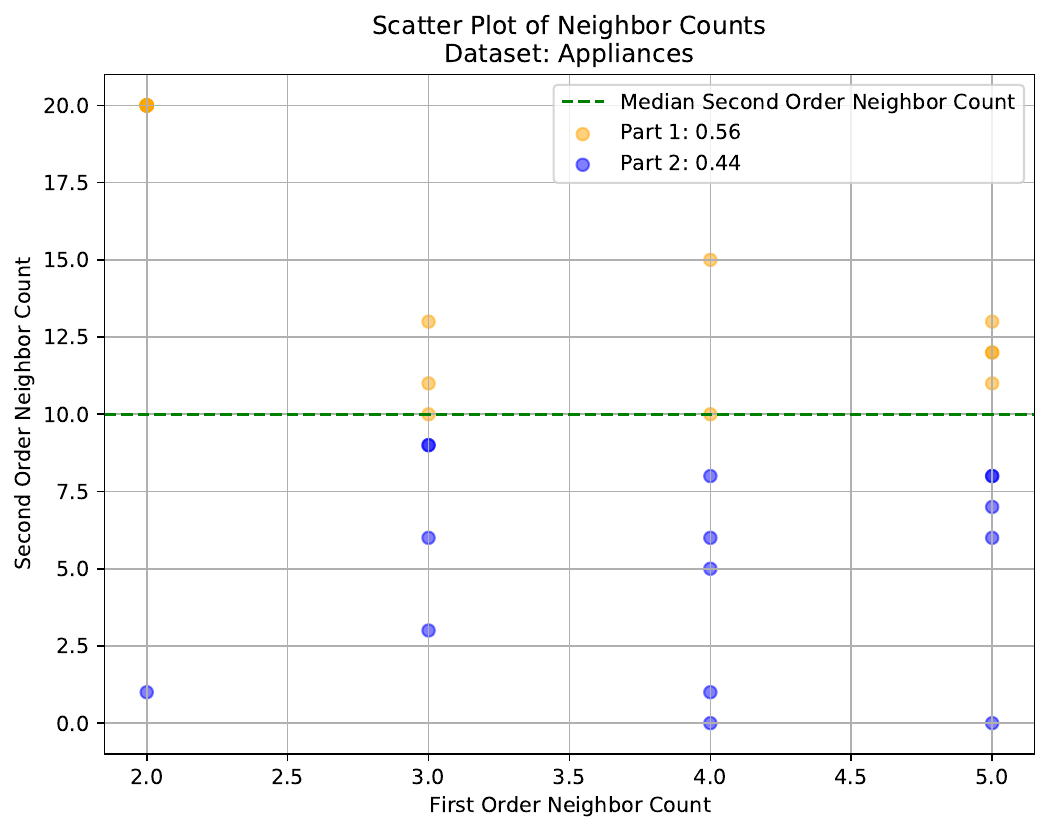}
    \caption{Appliances\_Long-tail\&Extreme\_Split}
    \label{fig:sub1}
  \end{subfigure}
  \hfill
  \begin{subfigure}[b]{0.45\textwidth}
    \includegraphics[width=\textwidth]{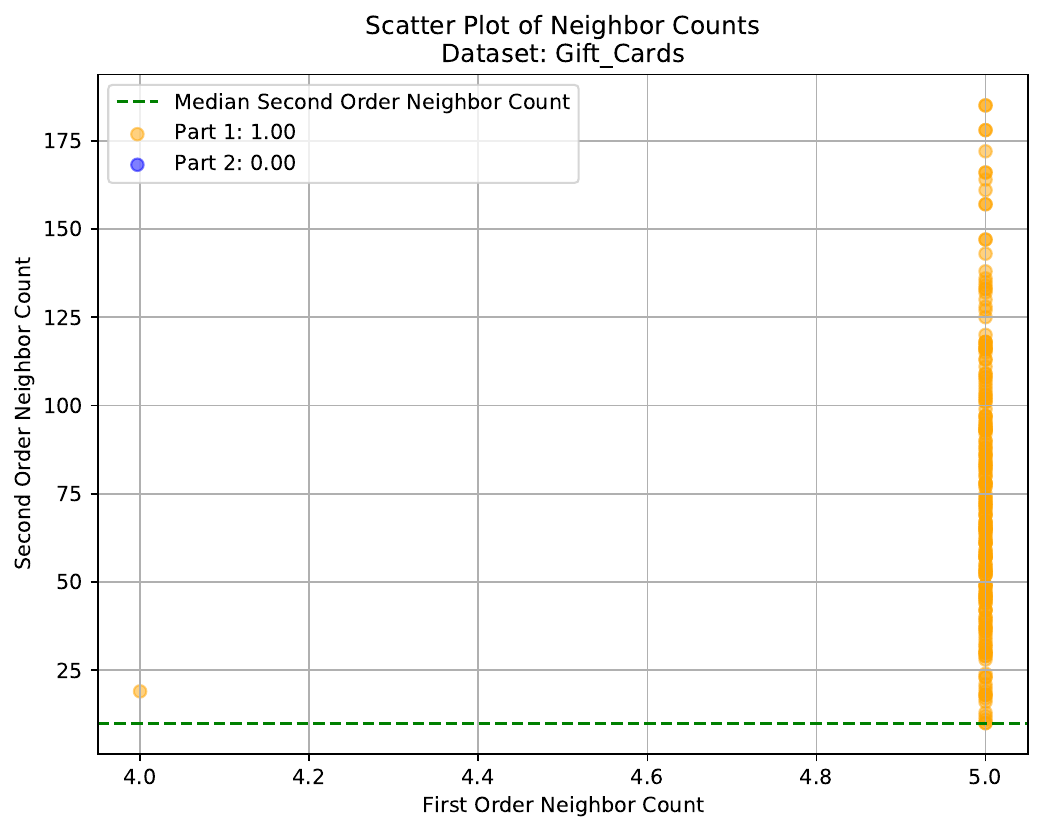}
    \caption{Gift\_Cards\_Long-tail\&Extreme\_Split}
    \label{fig:sub4}
  \end{subfigure}
  \vskip\baselineskip 
  \begin{subfigure}[b]{0.45\textwidth}
    \centering
    \includegraphics[width=\textwidth]{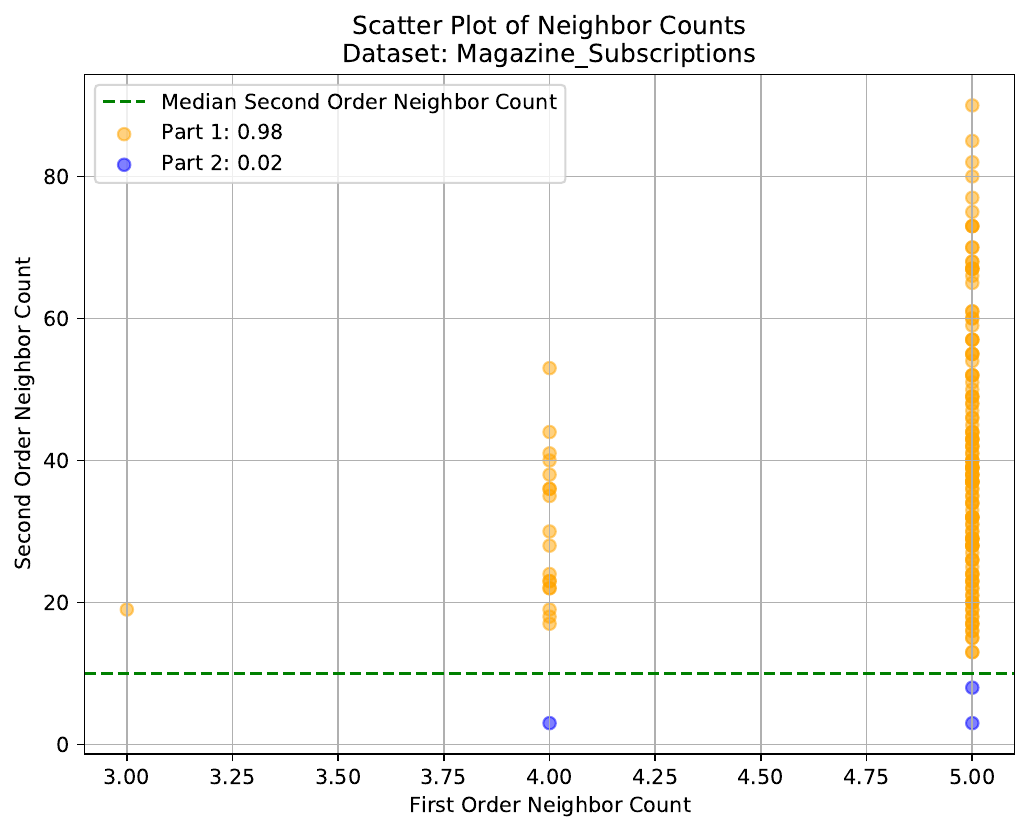}
    \caption{Magazine\_Subscriptions\_Long-tail\&Extreme\_Split}
    \label{fig:sub5}
  \end{subfigure}
  \caption{Data-Sparse User Division and Corresponding Proportions. The yellow points exhibit abundant second-order homogeneous relationships and are defined as long-tail users, while the blue points have sparse second-order homogeneous relationships and are defined as extreme cases.}
  \label{fig:sparse_division}
\end{figure*}

\noindent\textbf{Long-tail \& Extreme user split.} 
Figure~\ref{fig:sparse_division} illustrates a further division of sparse users due to data scarcity, including dividing situations and corresponding proportions. The lower region represents sparse users with limited self-data and few second-order neighbors, categorized as \textit{Extreme Situation}. The upper region represents sparse users with limited self-data but many second-order neighbors, which can be supplemented with synthesized data through second-order information, categorized as \textit{Long-tail Users}.

\begin{figure*}[htbp]
  \centering
  \begin{subfigure}[b]{0.43\textwidth}
    \includegraphics[width=\textwidth]{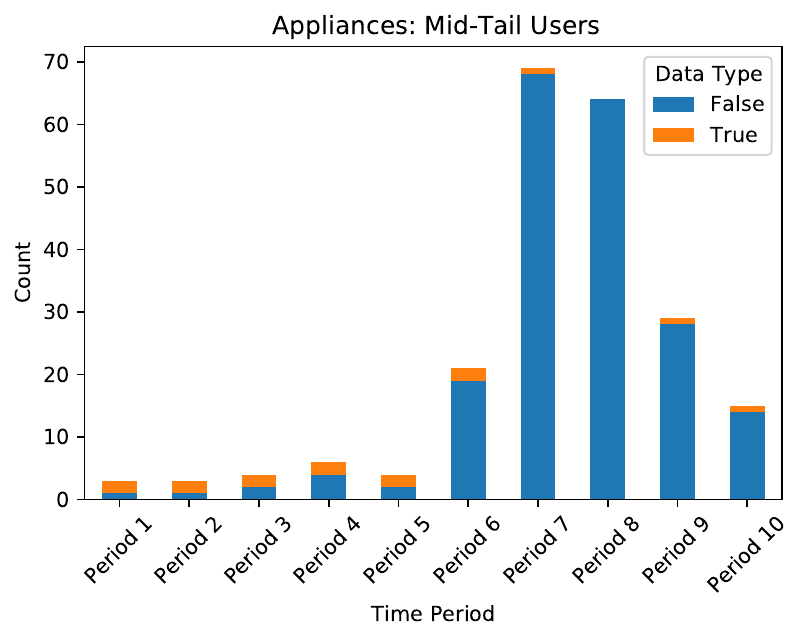}
    \caption{Appliances\_Mid-Tail}
    \label{fig:sub1}
  \end{subfigure}
  \hfill
  \begin{subfigure}[b]{0.43\textwidth}
    \includegraphics[width=\textwidth]{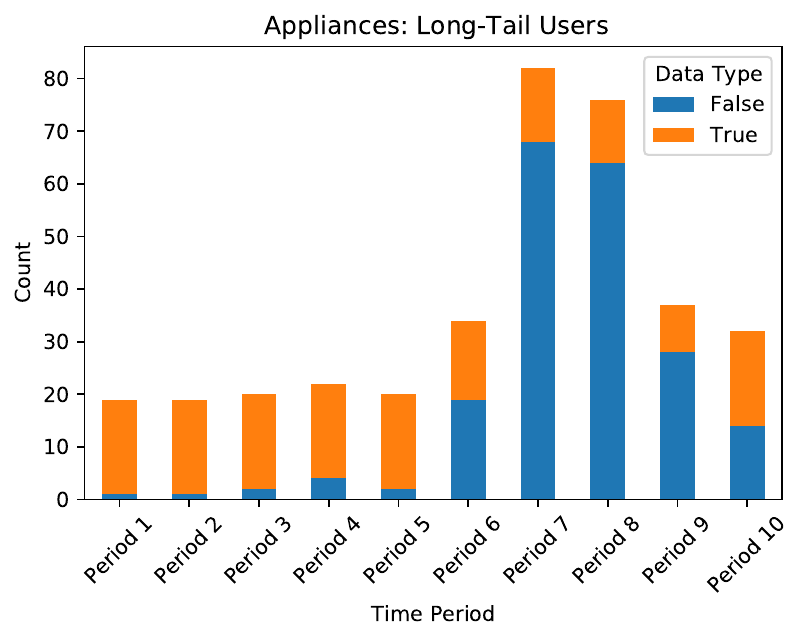}
    \caption{Appliances\_Long-Tail}
    \label{fig:sub4}
  \end{subfigure}
  \vskip\baselineskip 
\begin{subfigure}[b]{0.43\textwidth}
    \includegraphics[width=\textwidth]{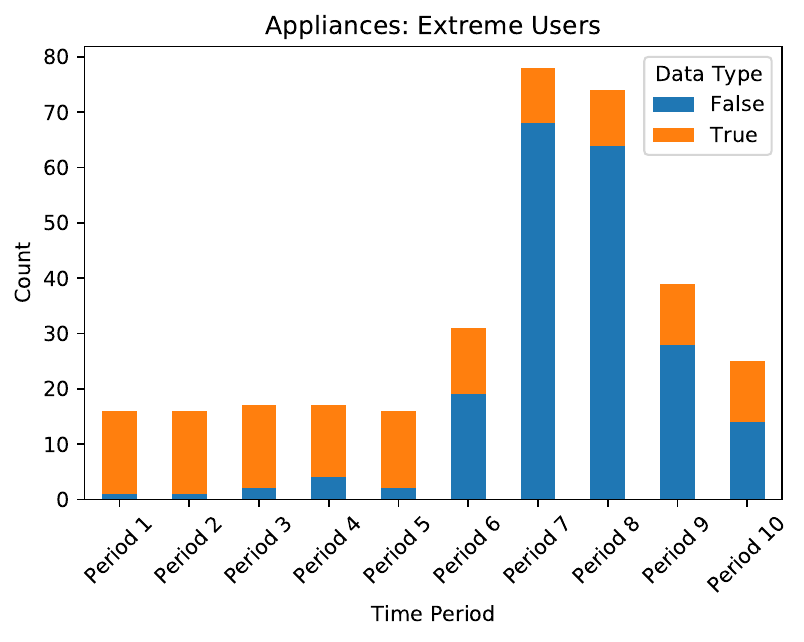}
    \caption{Appliances\_Extreme}
    \label{fig:sub1}
  \end{subfigure}
  \hfill
  \begin{subfigure}[b]{0.43\textwidth}
    \includegraphics[width=\textwidth]{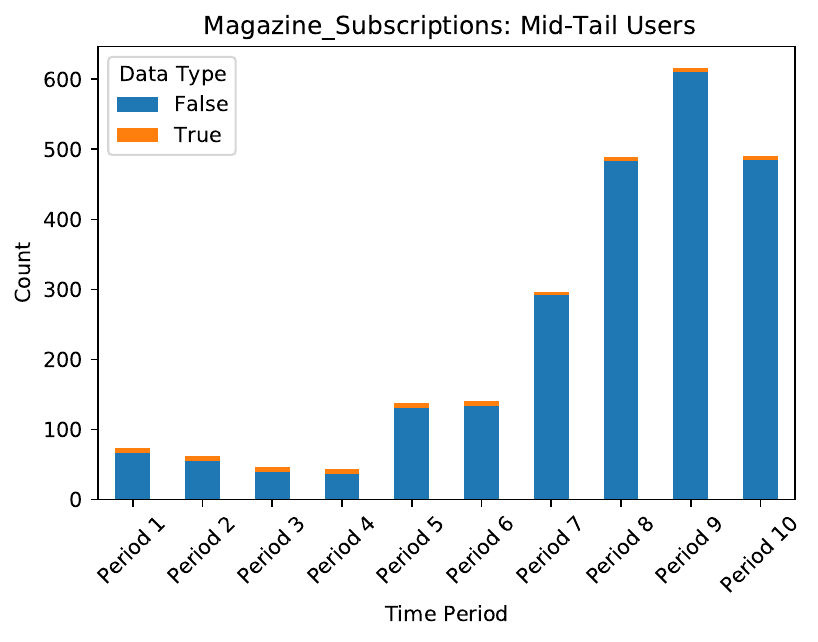}
    \caption{Magazine\_Subscriptions\_Mid-Tail}
    \label{fig:sub4}
  \end{subfigure}
    \vskip\baselineskip 
\begin{subfigure}[b]{0.43\textwidth}
    \includegraphics[width=\textwidth]{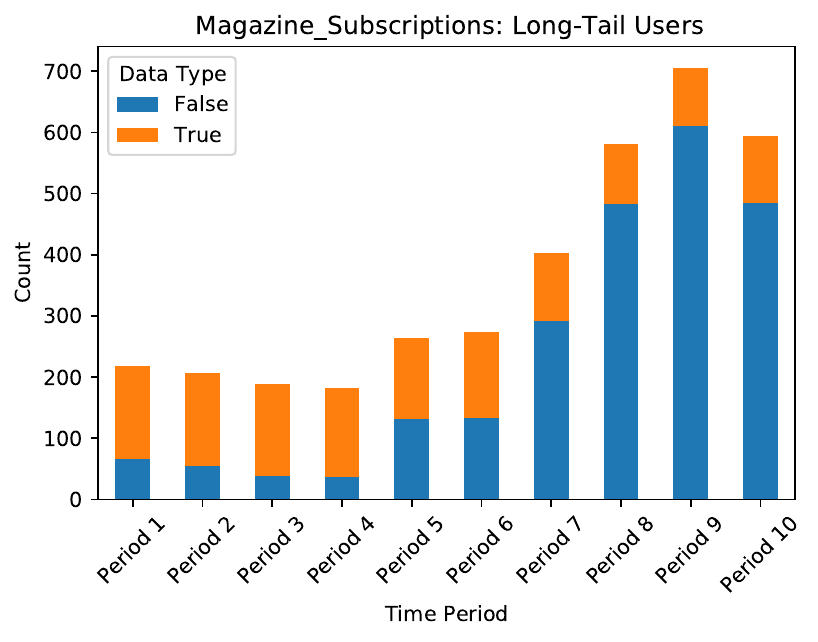}
    \caption{Magazine\_Subscriptions\_Long-Tail}
    \label{fig:sub1}
  \end{subfigure}
  \hfill
  \begin{subfigure}[b]{0.43\textwidth}
    \includegraphics[width=\textwidth]{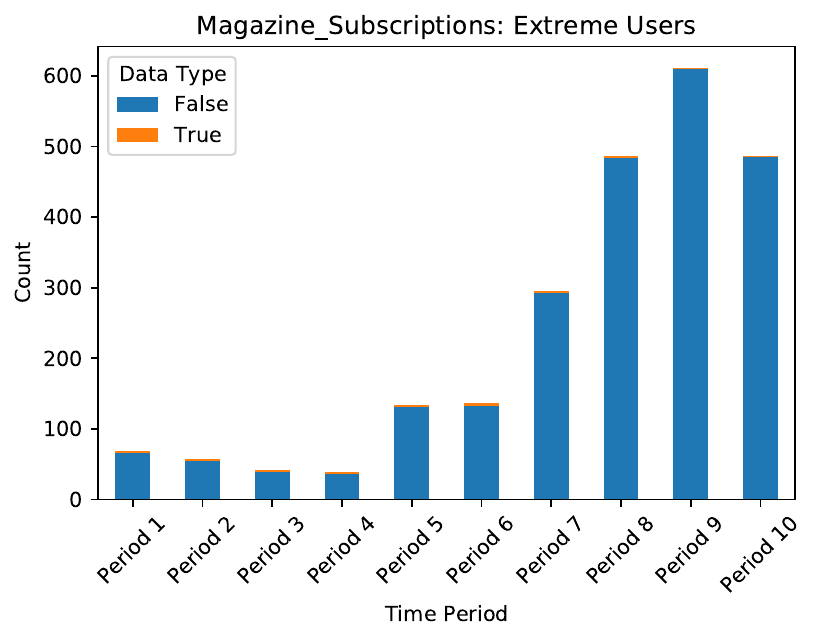}
    \caption{Magazine\_Subscriptions\_Extreme}
    \label{fig:sub4}
  \end{subfigure}
   \vskip\baselineskip 
\begin{subfigure}[b]{0.43\textwidth}
    \includegraphics[width=\textwidth]{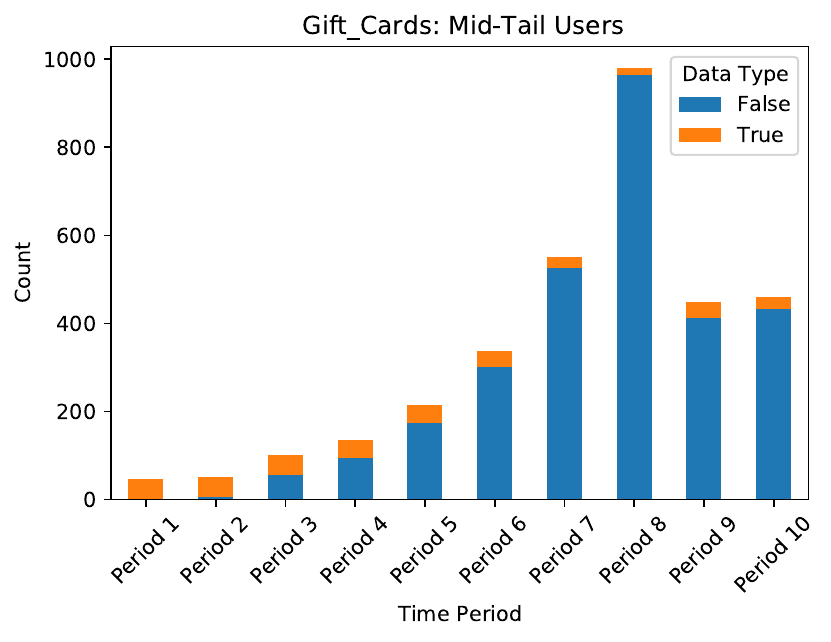}
    \caption{Gift\_Cards\_Mid-Tail}
    \label{fig:sub1}
  \end{subfigure}
  \hfill
  \begin{subfigure}[b]{0.43\textwidth}
    \includegraphics[width=\textwidth]{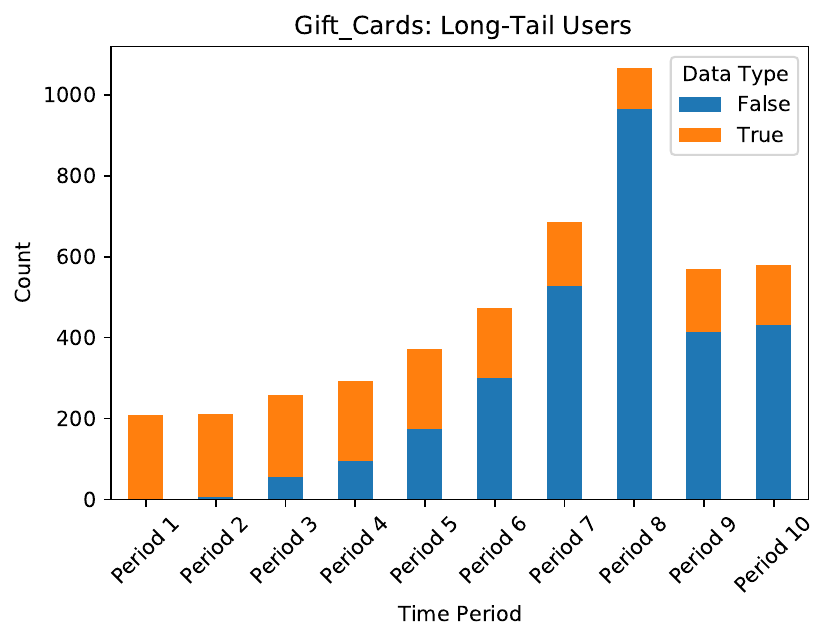}
    \caption{Gift\_Cards\_Long-Tail}
    \label{fig:sub4}
  \end{subfigure}
  \caption{Distribution of interpolation positions along the timeline corresponding to different sparse categories across datasets.}
  \label{fig:interpolated_distribution}
\end{figure*}

\noindent\textbf{Temporal distribution of interpolated data. }When performing data interpolation, it is necessary to determine the interpolation positions to use data synthesis methods for data synthesis, and then insert the synthesized data into the positions where interpolation is needed. Figure~\ref{fig:interpolated_distribution} shows the interpolation distribution of all types on all data over 10 time intervals.

\section{Prompts Templates}
\label{sec:prompt}
We utilize the official OpenAPI with the gpt-3.5-turbo\footnote{\url{https://platform.openai.com/docs/api-reference/models}} model for data synthesis. This section presents the prompts used for mid-tail, long-tail, and extreme scenarios, along with examples of profile generation and data synthesis by GPT.

\begin{figure*}[htbp]
\centering
\includegraphics[width=0.65\textwidth]{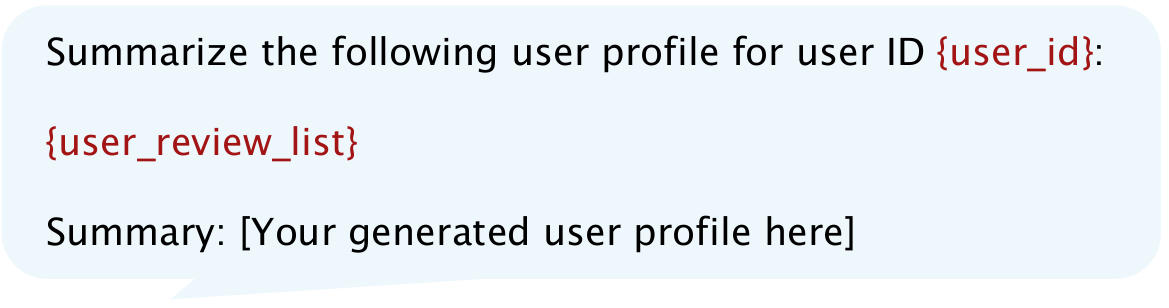}
\caption{The prompt used for generating user profiles in the mid-tail and extreme scenarios, defined as $\mathrm{P}_{um}$ and $\mathrm{P}_{ue}$ in the paper, respectively, takes as input the selected reviews of the user.}
\label{fig:P_ue}
\end{figure*}

\begin{figure*}[htbp]
\centering
\includegraphics[width=0.75\textwidth]{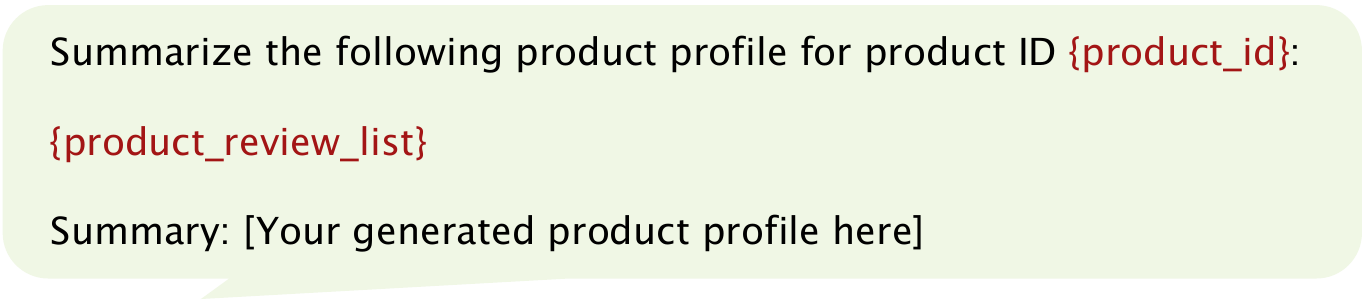}
\caption{The prompt used for generating product profiles in the mid-tail, long-tail, and extreme scenarios, defined as $\mathrm{P}_{pm}$, $\mathrm{P}_{pl}$, and $\mathrm{P}_{pe}$ in the paper, respectively, takes as input the selected reviews of the product.}
\label{fig:P_pe}
\end{figure*}

\begin{figure*}[htbp]
\centering
\includegraphics[width=0.9\textwidth]{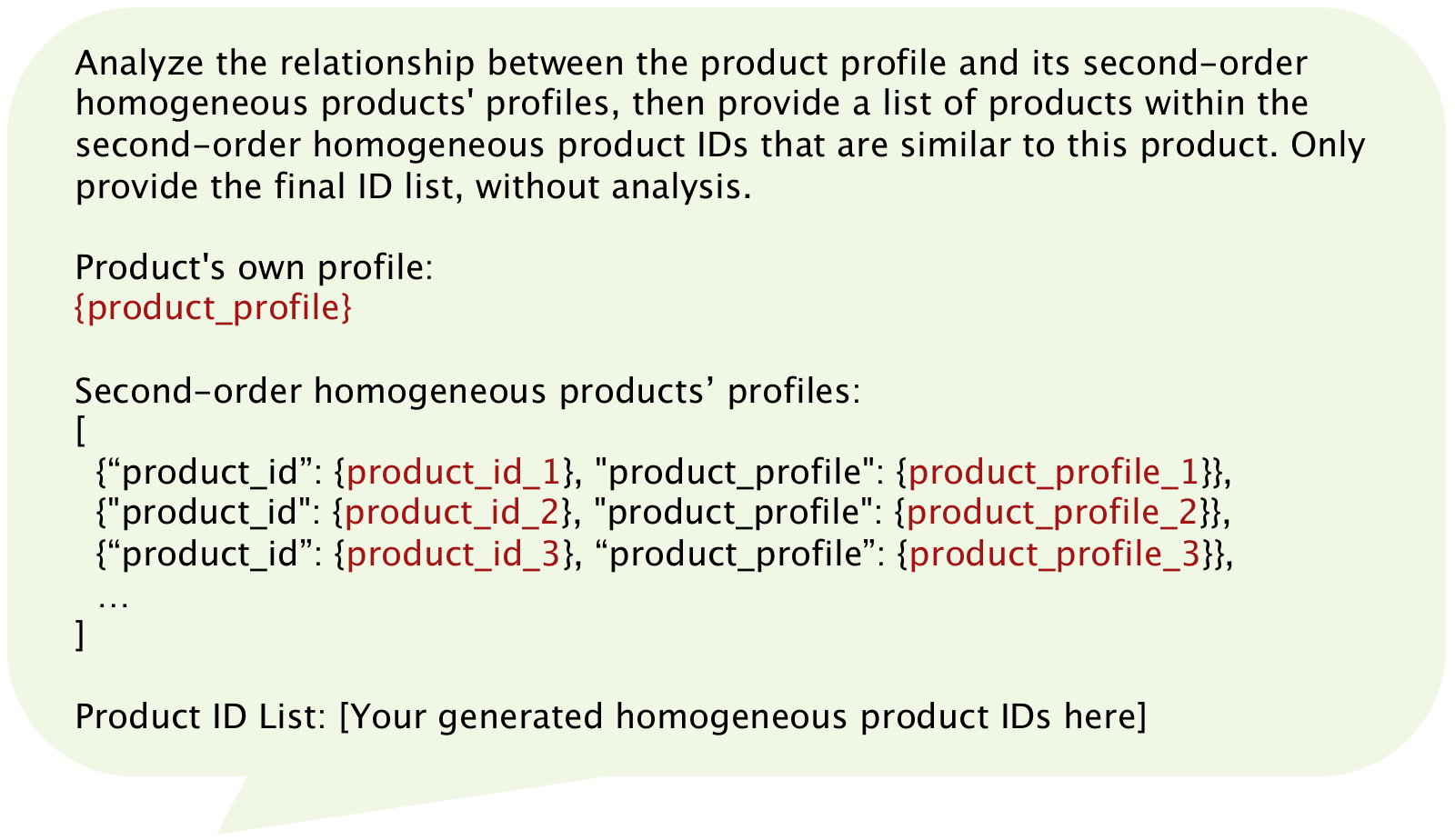}
\caption{The prompt used for selecting second-order homogeneous products in the mid-tail and long-tail scenarios, defined as $\mathrm{P}_{so}$ in the paper, takes as input the profile of the product itself along with the profile of the second-order homogeneous products.}
\label{fig:P_so}
\end{figure*}

\begin{figure*}[htbp]
\centering
\includegraphics[width=0.9\textwidth]{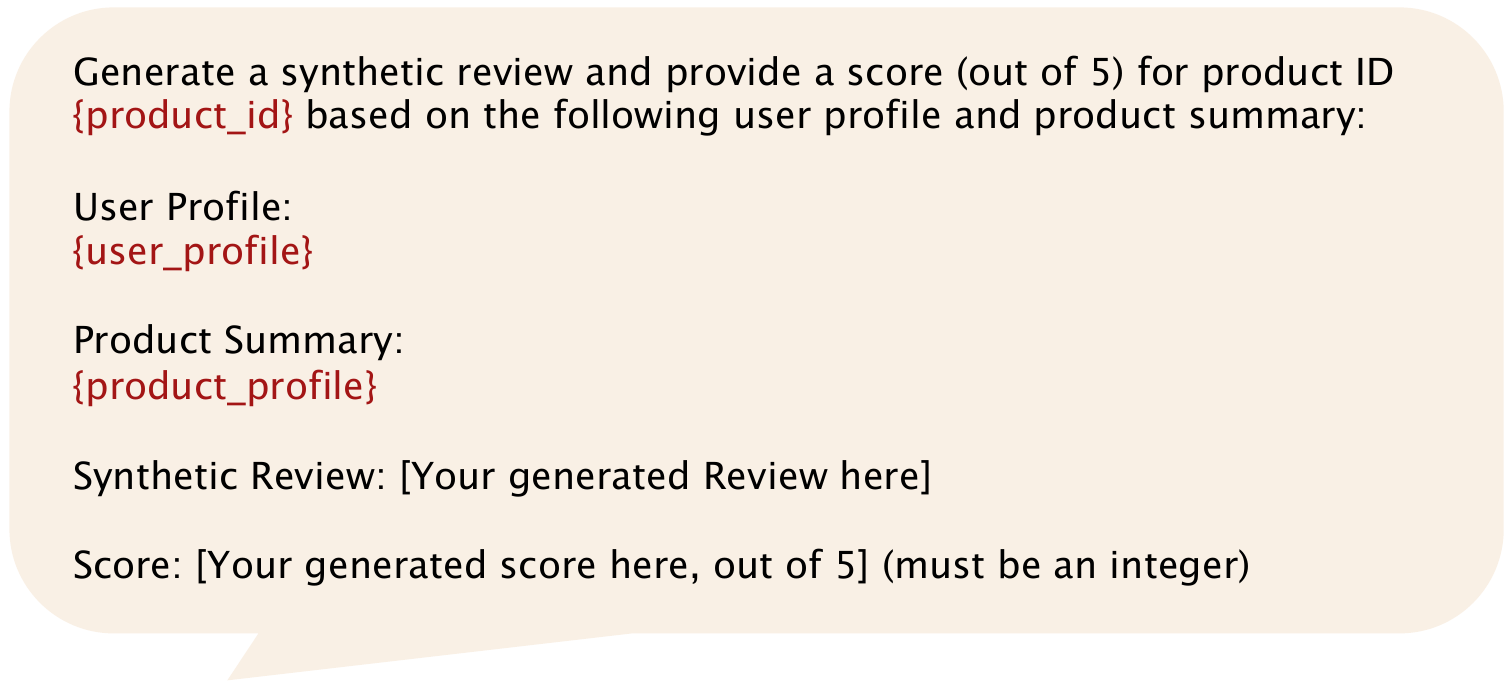}
\caption{The prompt used for synthesizing data for mid-tail, long-tail, and extreme user scenarios, defined as $\mathrm{P}_{sd}$ in the paper, takes as input the user profile and the product profile.}
\label{fig:P_ds}
\end{figure*}

In the mid-tail scenario, $\mathrm{P}_{um}$ in Figure~\ref{fig:P_ue} is used for generating user profiles, $\mathrm{P}_{pm}$ in Figure~\ref{fig:P_pe} for generating product profiles, $\mathrm{P}_{so}$ in Figure~\ref{fig:P_so} for selecting second-order homogeneous products, and $\mathrm{P}_{sd}$ in Figure~\ref{fig:P_ds} for data synthesis.

\begin{figure*}[htbp]
\centering
\includegraphics[width=0.9\textwidth]{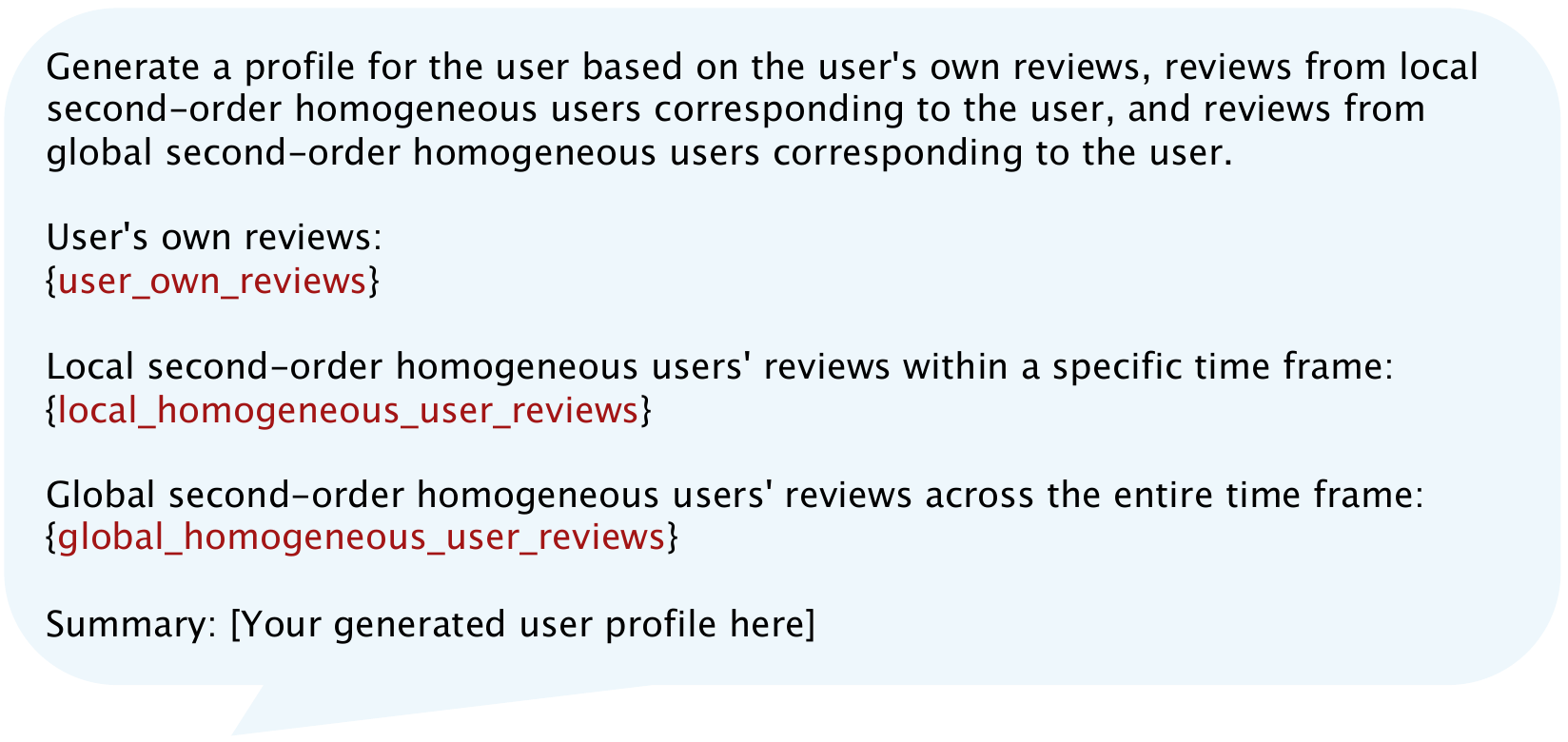}
\caption{The prompt used for understanding the local and global user relationships in the long-tail scenario, defined as $\mathrm{P}_{ul}$ in the paper, takes as input the user's own reviews, the reviews of locally second-order homogeneous users, and the reviews of globally second-order homogeneous users.}
\label{fig:P_ul}
\end{figure*}

In the long-tail scenario, $\mathrm{P}_{ul}$ is used for generating user profiles, as shown in Figure~\ref{fig:P_ul}. $\mathrm{P}_{pl}=\mathrm{P}_{pm}$ is used for generating product profiles. Additionally, $\mathrm{P}_{so}$ and $\mathrm{P}_{sd}$ remain the same as in the mid-tail scenario.

In the extreme scenario, $\mathrm{P}_{ue}=\mathrm{P}_{um}$ is used for generating user profiles, $\mathrm{P}_{pe}=\mathrm{P}_{pm}$ is used for generating product profiles, and $\mathrm{P}_{sd}$ remains the same as in the mid-tail scenario.

\begin{figure*}[htbp]
\centering
\includegraphics[width=0.8\textwidth]{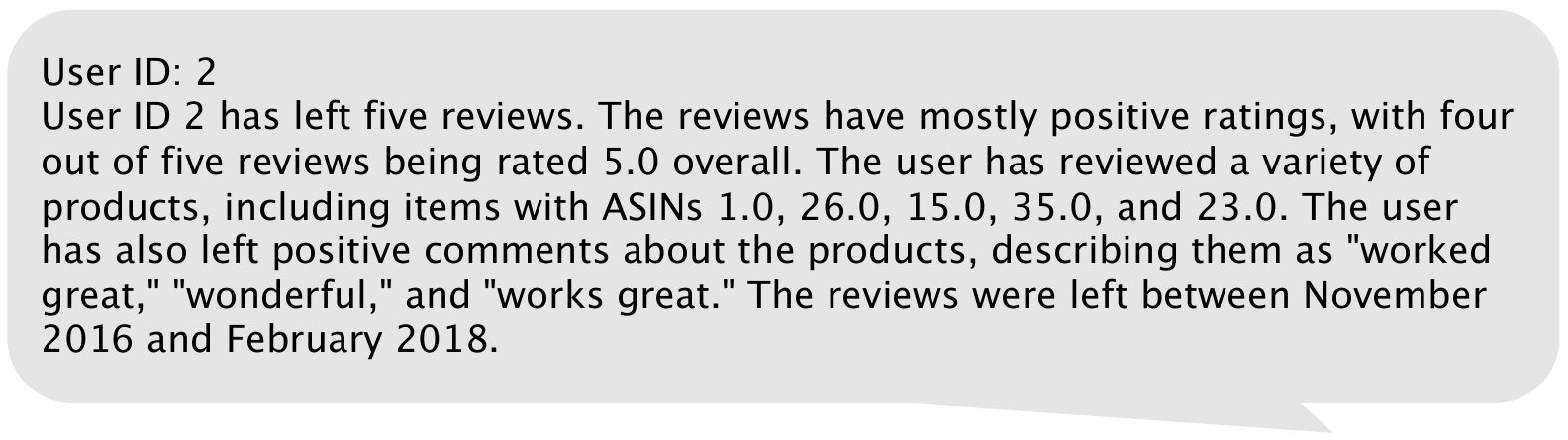}
\caption{Example of user profiles generated by GPT.}
\label{fig:U_exp}
\end{figure*}

\begin{figure*}[htbp]
\centering
\includegraphics[width=0.8\textwidth]{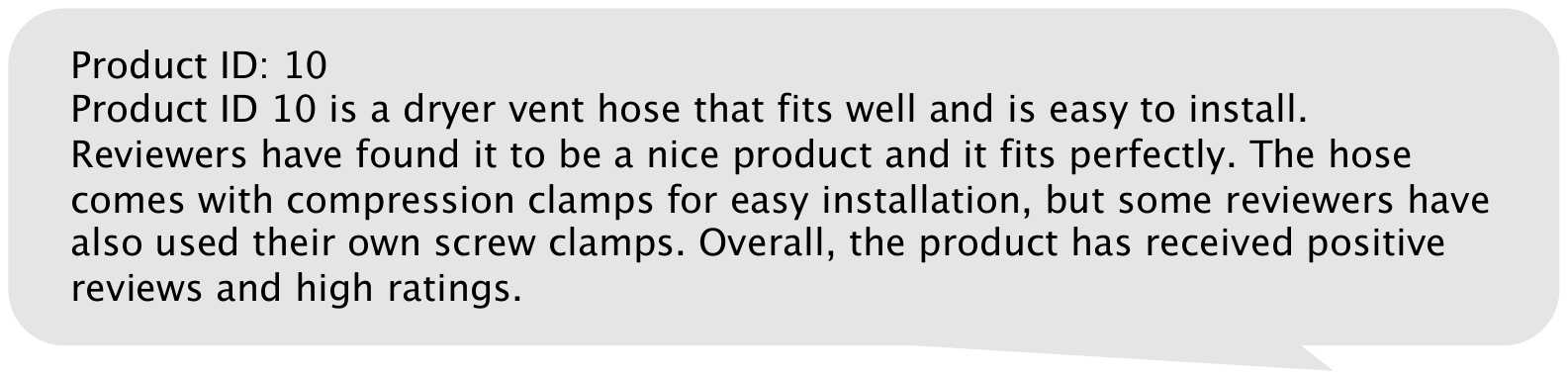}
\caption{Example of product profiles generated by GPT.}
\label{fig:P_exp}
\end{figure*}

\begin{figure*}[htbp]
\centering
\includegraphics[width=0.9\textwidth]{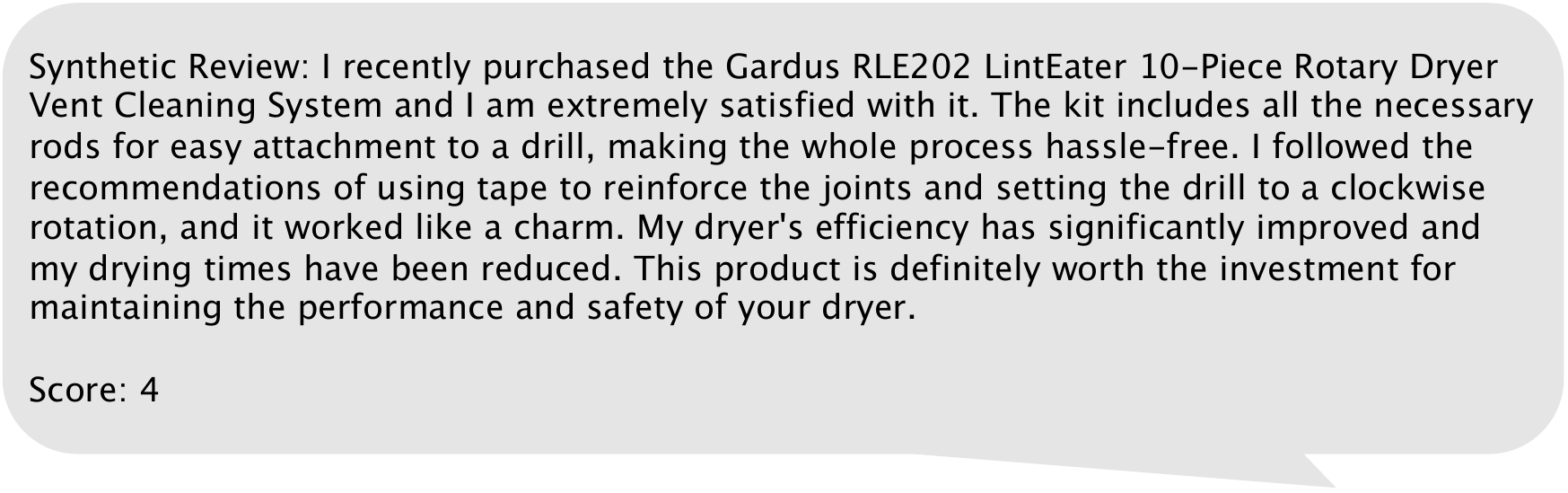}
\caption{Example of synthetic data demonstrating positive sentiment generated by GPT.}
\label{fig:sd_exp1}
\end{figure*}

\begin{figure*}[htbp]
\centering
\includegraphics[width=0.9\textwidth]{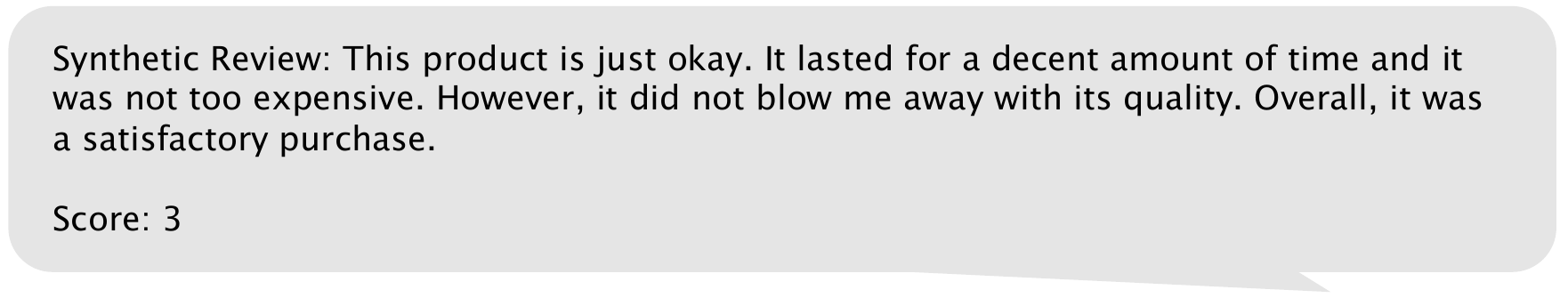}
\caption{Example of synthetic data demonstrating neutral sentiment generated by GPT.}
\label{fig:sd_exp2}
\end{figure*}

Figure~\ref{fig:U_exp} and Figure~\ref{fig:P_exp} respectively illustrate an example of a user profile and a product profile generated by GPT. Figure~\ref{fig:sd_exp1} demonstrates an example of synthesized data with a positive sentiment, while Figure~\ref{fig:sd_exp2} shows an example of synthesized data with a neutral sentiment.

\begin{table*}[htbp]
    \centering
    \resizebox{1.8\columnwidth}{!}{
\begin{tabular}{|l|c|c|c|c|c|}
\hline
\textbf{Dataset}          & \textbf{Total R} & \textbf{U10 R} & \textbf{U10 Rproportion} & \textbf{U5 R} & \textbf{U5 Rproportion}  \\ \hline
Magazine\_Subscriptions   & 2330                    & 1178                      & 0.506                        & 764                       & 0.328                        \\ \hline
Appliances                & 203                     & 87                        & 0.429                        & 116                       & 0.571                        \\ \hline
Gift\_Cards               & 2966                    & 1502                      & 0.506                        & 1044                      & 0.352                        \\ \hline
    \end{tabular}}
\caption{
Statistical analysis of the ratio of user-associated reviews to the total review count across various hierarchical levels. U10 R refers to the number of reviews associated with users with ten or fewer reviews. U5 R refers to the number of reviews associated with users with five or fewer reviews.}
\label{tab:dif_user_statistic}
\end{table*}

\end{document}